\pdfoutput=1
\documentclass[11pt]{article}

\usepackage[final]{acl}

\definecolor{deepgreen}{RGB}{0, 70, 0}
\usepackage{amsmath}
\renewcommand{\sectionautorefname}{\S\@gobble}

\renewcommand{\subsectionautorefname}{\S\@gobble} 
\renewcommand{\subsubsectionautorefname}{\S\@gobble}
\renewcommand{\appendixautorefname}{\S\@gobble}

\usepackage{longtable}
\usepackage{geometry}
\usepackage{array}
\usepackage{booktabs}
\usepackage{ragged2e}
\usepackage{caption}

\makeatother

\definecolor{olmoeDarkYellow}{HTML}{fdac15}
\definecolor{defaultblue}{HTML}{0077B6}
\definecolor{defaultlightblue}{HTML}{00B4D8}
\definecolor{blue}{HTML}{03045E}
\definecolor{blueb}{HTML}{0077B6}
\definecolor{bluec}{HTML}{00B4D8}
\definecolor{blued}{HTML}{90E0EF}
\definecolor{bluee}{HTML}{CAF0F8}

\definecolor{olmoeDarkYellow}{HTML}{fdac15}
\definecolor{defaultblue}{HTML}{0077B6}
\definecolor{defaultlightblue}{HTML}{00B4D8}
\definecolor{blue}{HTML}{03045E}
\definecolor{blueb}{HTML}{0077B6}
\definecolor{bluec}{HTML}{00B4D8}
\definecolor{blued}{HTML}{90E0EF}
\definecolor{bluee}{HTML}{CAF0F8}

\usepackage{tcolorbox}
\tcbuselibrary{listings,breakable}

\usepackage{times}
\usepackage{latexsym}
\usepackage{booktabs}
\usepackage{multirow}
\usepackage{colortbl}
\usepackage{tabularx}
\usepackage[T1]{fontenc}
\usepackage[utf8]{inputenc}

\usepackage{microtype}

\usepackage{inconsolata}

\usepackage{graphicx}

\title{\raisebox{-0.3cm}{\includegraphics[width=1.2cm,height=1.2cm,keepaspectratio]{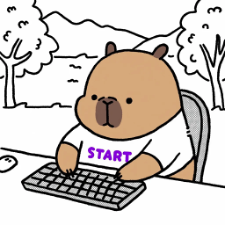}}START: Self-taught Reasoner with Tools}

\author{Chengpeng Li\textsuperscript{1,2}\thanks{Work done during internships at Alibaba Group.}, 
Mingfeng Xue\textsuperscript{2$^*$}, 
Zhenru Zhang\textsuperscript{2}, 
Jiaxi Yang\textsuperscript{2}$^*$, 
Beichen Zhang\textsuperscript{2}$^*$, \\ 
\textbf{Xiang Wang\textsuperscript{1}, 
Bowen Yu\textsuperscript{2}, 
Binyuan Hui\textsuperscript{2}, 
Junyang Lin\textsuperscript{2}, 
Dayiheng Liu\textsuperscript{2}\thanks{Corresponding author}} \\ 
\textsuperscript{1}University of Science and Technology of China \\ 
\textsuperscript{2}Alibaba Group \\ 
\texttt{\{lichengpeng.lcp,liudayiheng.ldyh\}@alibaba-inc.com} 
}

\begin{document}
\maketitle
\begin{abstract}

Large reasoning models (LRMs) like OpenAI-o1 and DeepSeek-R1 have demonstrated remarkable capabilities in complex reasoning tasks through the utilization of long Chain-of-thought (CoT). However, these models often suffer from hallucinations and inefficiencies due to their reliance solely on internal reasoning processes. In this paper, we introduce START (\textbf{S}elf-\textbf{Ta}ught \textbf{R}easoner with \textbf{T}ools), a novel tool-integrated long CoT reasoning LLM that significantly enhances reasoning capabilities by leveraging external tools. Through code execution, START is capable of performing complex computations , self-checking, exploring diverse methods, and self-debugging, thereby addressing  limitations of LRMs.
% A key innovation in START is the Hint-infer technique. We demonstrate that inserting artificially designed hints (e.g., “Wait, maybe using Python here is a good idea.”) during the inference process of LRMs effectively stimulates the model's ability to utilize external tools without the need for demonstration data.  This self-learning framework, which includes Hint-RFT and Reject Sampling Fine-Tuning, facilitates autonomous data generation and self-improvement, significantly enhancing performance and efficient tool utilization with only one input LRM required. 
The core innovation of START lies in its self-learning framework, which comprises two key techniques: 1) Hint-infer: We demonstrate that inserting artificially designed hints 
(e.g., “Wait, maybe using Python here is a good idea.”) 
during the inference process of a LRM effectively stimulates its ability to utilize external tools without the need for any demonstration data. Hint-infer can also serve as a simple and effective sequential test-time scaling method; 2) Hint Rejection Sampling Fine-Tuning (Hint-RFT): Hint-RFT combines Hint-infer and RFT by scoring, filtering, and modifying the reasoning trajectories with tool invocation generated by a LRM via Hint-infer, followed by fine-tuning the LRM.
Through this framework, we have fine-tuned the QwQ-32B-Preview model to achieve the START.
% Empirical evaluations on a suite of challenging benchmarks confirm START's superior performance. 
% On PhD-level science QA (GPQA), competition-level math benchmarks(AMC23, AIME24, AIME25) and competition-level code benchmark(LiveCodeBench), START significantly outperforms the base QwQ-32B and achieves performance comparable to state-of-the-art open-weights R1-Distil-Qwen-32B and proprietary o1-Preview.
On PhD-level science QA (GPQA), competition-level math benchmarks (AMC23, AIME24, AIME25), and the competition-level code benchmark (LiveCodeBench), START achieves accuracy rates of 63.6\%, 95.0\%, 66.7\%, 47.1\%, and 47.3\%, respectively. It significantly outperforms the base QwQ-32B-Preview and achieves performance comparable to the state-of-the-art open-weight model R1-Distill-Qwen-32B and the proprietary model o1-Preview.

% to generate tool-assisted reasoning data by embedding 

% within QWQ's inference process on a seed dataset. Interestingly, we do not teach the model how to invoke the tool, but simply give a hint that will stimulate the model's potential tool invocation capabilities. Subsequently, through iterative Hint-RFT on a larger dataset, we expand the training dataset to include multiple tool invocation scenarios, culminating in the START-Instruct model. 

% Furthermore, we develop START-RM, an outcome-based generative reward model trained on correct and wrong tool-integrated trajectories from iterative RFT and Hint-RFT. START-RM can significantly improve the best-of-N performance of START during the inference stage. 

\end{abstract}

\section{Introduction}
\label{sec:intro}

% The advent of large language models (LLMs) has revolutionized natural language processing, enabling significant advancements in a myriad of applications, from text generation to complex reasoning tasks~\citep{gpt4,qwen2,gemini}. A pivotal innovation that underpins this progress is the concept of long chains of thought (CoT), which emulates human-like reasoning by allowing models to generate extended internal deliberations before arriving at a conclusion ~\citep{CoT,o1}. Despite their successes, existing LLMs employing long CoT may grapple with issues such as hallucinations and inefficiencies, primarily stemming from their exclusive dependence on internal reasoning mechanisms.

% To address these limitations, tool-integrated reasoning(TIR) emerges as a promising strategy~\citep{toolformer,toollearning,tora}. Tool integration allows LLMs to leverage specialized functionalities, such as the execution of code or the access to real-time data, to improve the efficiency and reliability of their reasoning. For instance, invoking a Python interpreter can enable a model to verify mathematical solutions or debug generated code, thereby mitigating errors and reducing hallucinations. In this paper, we only focus on Python interpreter, since it is important and representative for many reasoning tasks.

The evolution of reasoning capabilities in large language models (LLMs) has followed a paradigm shift marked by increasingly sophisticated thinking patterns. The chain-of-thought (CoT) approach~\citep{CoT} pioneers this progression by enabling models to decompose problems through explicit intermediate reasoning steps. Then, the breakthrough comes with reinforcement learning  exemplified by OpenAI-o1~\citep{o1} and DeepSeek-R1~\citep{DeepSeekR1}, establishing a new paradigm termed long CoT, which emulates human-like cognitive strategies including self-refine, self-reflection, multi-strategy exploration and so on. Despite these advancements, fundamental limitations persist in long Chain-of-thought (CoT) approaches, such as hallucinations when facing complex computations or simulations, primarily due to their exclusive dependence on internal reasoning mechanisms~\citep{search-o1}. Tool-integrated reasoning (TIR)~\citep{tora}, another approach that improves traditional CoT through tool invocation(typically code interpreter), can effectively mitigate the issues that arise in Long CoT. OpenAI o1~\citep{o1} reported that they trained o1 to use external tools, especially for writing and executing code in a secure environment, but they did not provide any technical details on how this is achieved. Therefore, a straightforward yet important question arises: \textit{How can we synergistically combine long CoT with TIR?}

\begin{figure*}[ht]
    \centering
    \includegraphics[width=1.0\linewidth]{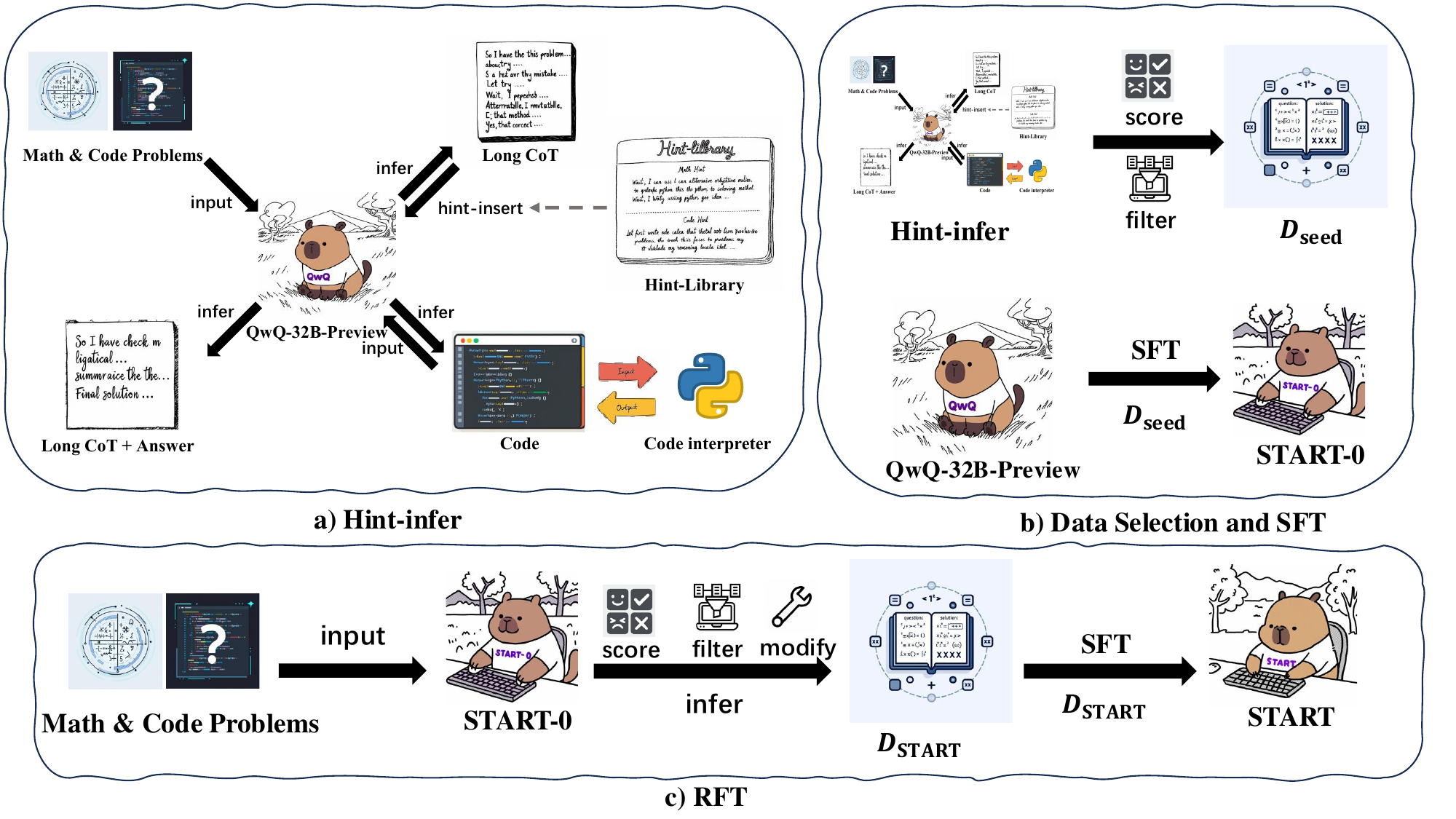}
    \caption{\textbf{Training framework for START.} Training Framework for START. START's training involves two phases: Hint-RFT followed by RFT. a) Hint-infer: code/math data is processed by QwQ-32B-Preview, with responses truncated at predefined terminators. Context-aware hints from a Hint-Library are injected at truncation points (including endpoints), and QwQ-32B-Preview resumes inference using a code interpreter for Python execution feedback. b) Hint-RFT: Hint-infer outputs undergo rule-based scoring, filtering, and content modification to create $D_{\text{seed}}$
 . QwQ-32B-Preview is then fine-tuned on $D_{\text{seed}}$
  to produce START-0, enabling self-aware tool usage. c) RFT: START-0 generates self-distilled trajectories to build $D_{\text{START}}$
  (enhancing diversity/tool-use patterns), followed by fine-tuning to produce START.}
    \label{fig:mainfig}
\end{figure*}

\begin{figure*}[ht]
    \centering
    \includegraphics[width=1.05\linewidth]{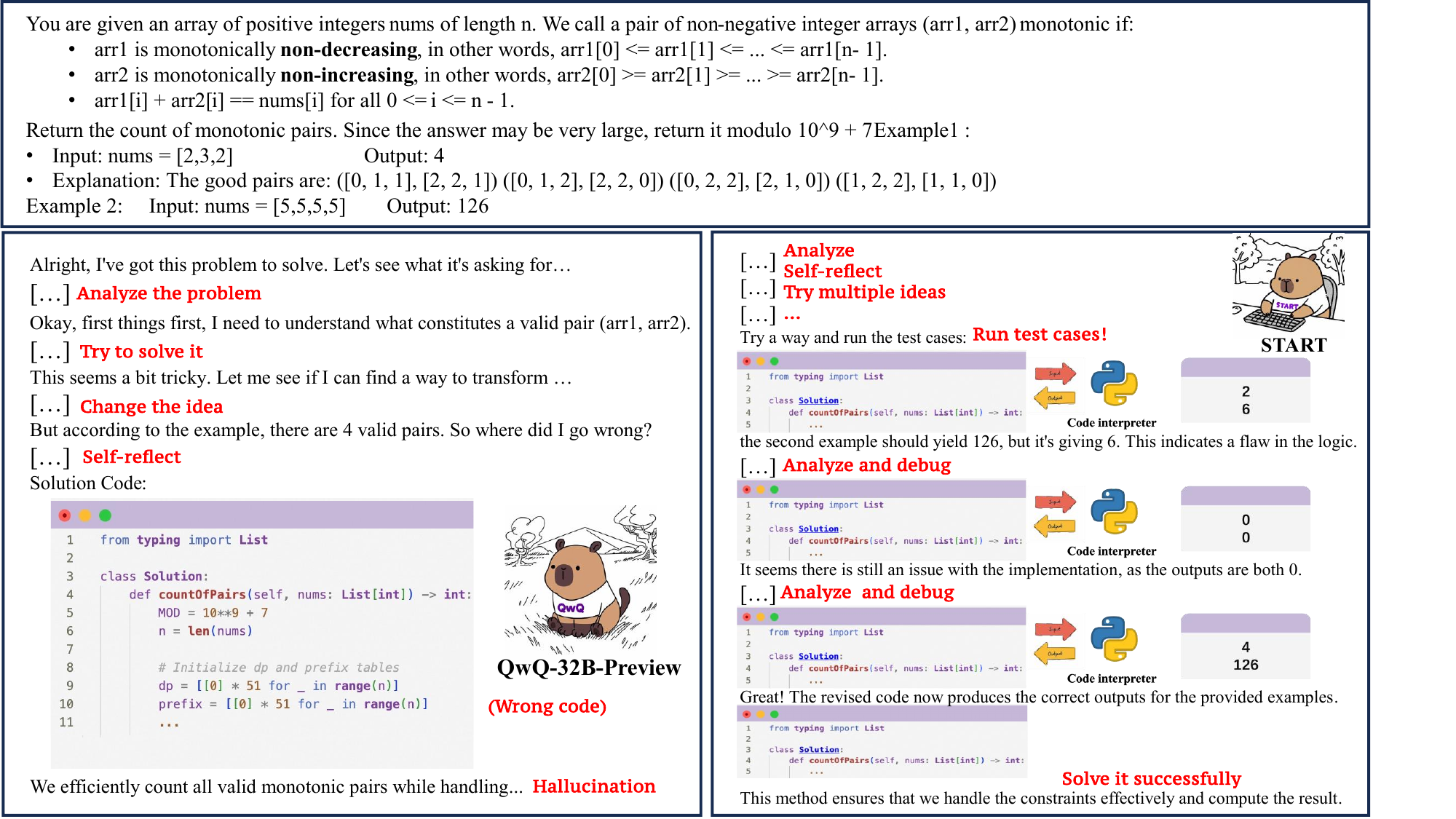}
    \caption{Comparison between the responses generated by QwQ-32B-Preview and START. This is a question from LiveCodeBench with a difficulty level of "hard". QwQ-32B-Preview employs long-chain CoT with self-reflection and trying different approaches, yet hallucinates during complex test case analysis, leading to flawed solutions. START retains QwQ-32B-Preview's cognitive framework but integrates code execution: (1) Runs code via interpreter, (2) Detects output mismatch, (3) Iteratively analyzes and debugs, and (4) Gives the final solution. See more cases of START in Appendix ~\ref{casestudy}}
    \label{fig:maincase}
\end{figure*}

\begin{figure*}[ht]
    \centering
    \includegraphics[width=1.0\linewidth]{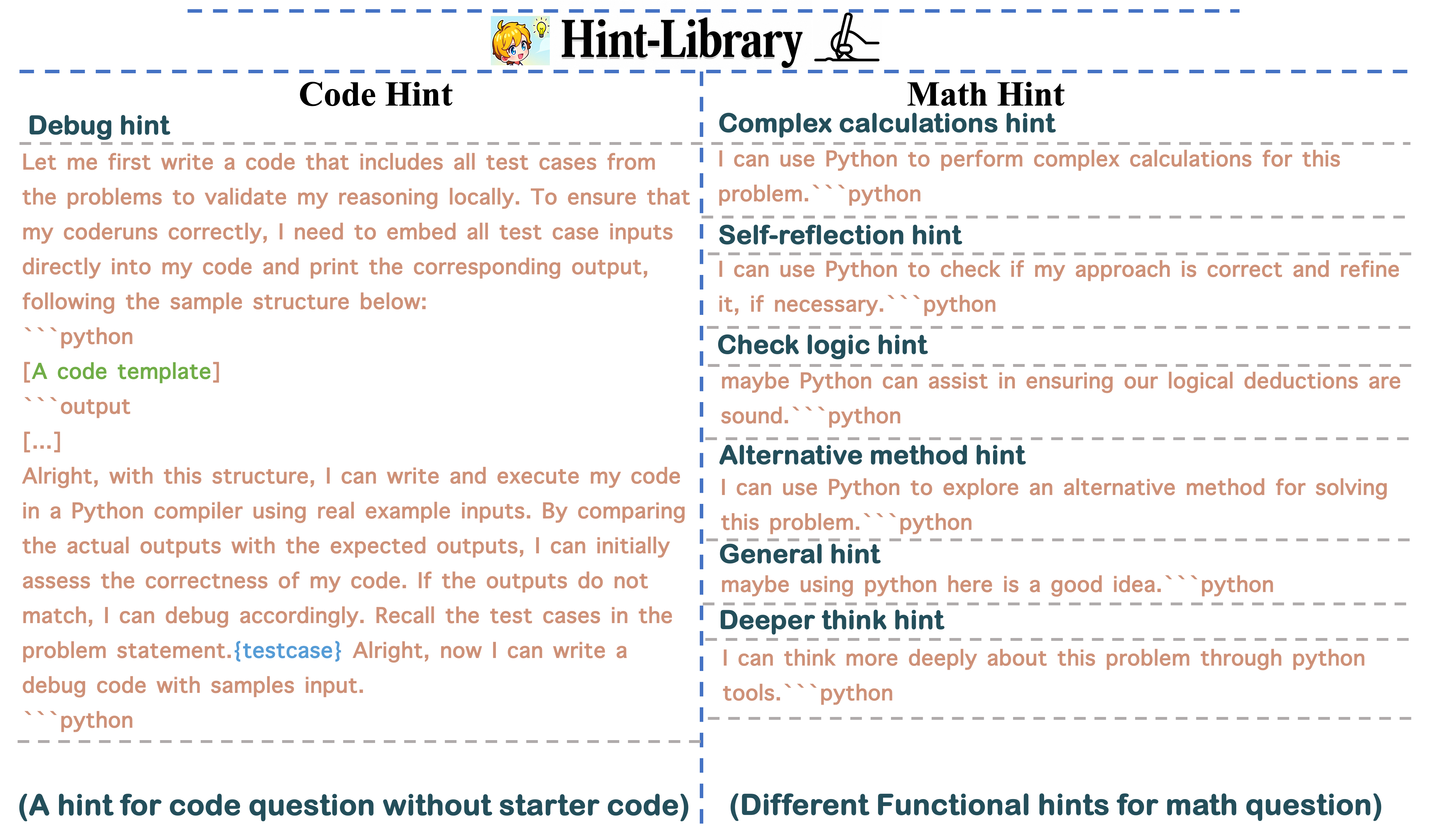}
    \caption{\textbf{Hint-Library}. Code generation tasks: Debug hint guides test case review and local code validation. The code template is in \ref{hint-infer-prompt}. Math reasoning: Domain-specific hints (e.g., Complex Calculations, Self-Reflection, Logic Check, Alternative Methods) steer code-aided reasoning behaviors.}
    \label{fig:hint-library}
\end{figure*}

% \begin{figure*}[h]
%     \centering
%     \includegraphics[width=1.0\linewidth]{start_format_3.pdf}
%     \caption{The data format comparision}
%     \label{fig:dataformat}
% \end{figure*}

In this paper, we focus exclusively on Python interpreter invocation, as it is both important and representative of many reasoning tasks~\citep{deepseekmath,qwen2.5math}. The fundamental challenge lies in synthesizing data that includes calls to a Python interpreter within Long Chain-of-thought (CoT). We have tried using direct prompt, well-designed prompt~\citep{search-o1}, and in-context prompt~\citep{tora,toolformer} on AIME24 and LivecodeBench with QwQ-32B-Preview~\citep{QwQ} and DeepSeek-R1, but none were successful in prompting the model to invoke the Python tool during the long CoT(see more in Appendix \ref{prompting method}). A possible reason is that large reasoning models(LRMs) typically focus solely on problem-solving during training for complex reasoning tasks, resulting in a loss of generalization in instruction following. Considering the nature of next-token prediction in LRMs, we attempt to insert some hints directly during or at the end of the LRMs' reasoning process, aiming to directly prompt the model to write code and invoke code interpreter. We are surprised to discover that LLMs indeed possess the corresponding capabilities. For mathematical tasks, simply inserting basic hints along with Python identifiers enables the LLM to follow the hints and write the appropriate code. In contrast, for coding generation tasks, carefully designed hints and code templates are necessary to activate the model's ability to execute candidate code on test cases on its own during the long CoT. We refer to the paradigm of LRM inference aided by hints as \textbf{Hint-infer}.

\begin{figure}[ht]
    \centering
    \includegraphics[width=1.0\linewidth]{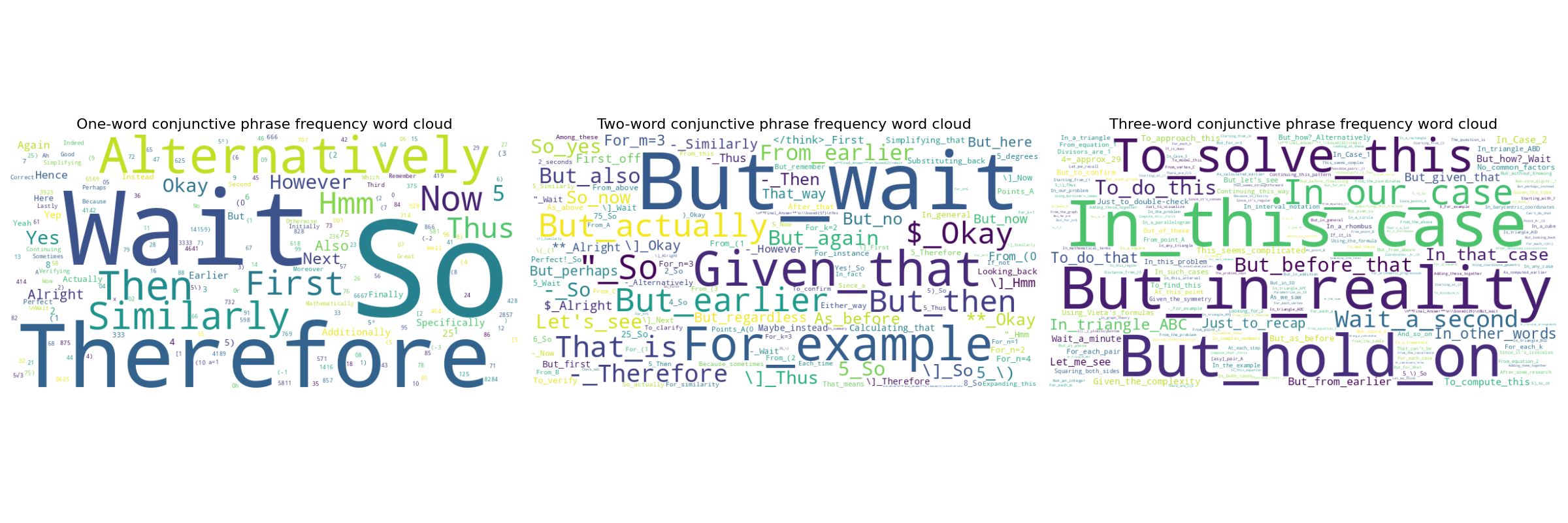}
    \caption{Word cloud of conjunction frequency statistics from QwQ-32B-Preview infering on $D_{\text{seed}}$.}
    \label{fig:Word cloud}
    % \vspace{-0.5}
\end{figure}

Based on above Hint-infer, we present START: \textbf{S}elf-\textbf{Ta}ught \textbf{R}easoner with \textbf{T}ools, a LRM that synergizes Long CoT and TIR, which we refer to as Long TIR. The whole traing framework is illustrated in Figure\ref{fig:mainfig}. First, we design a set of hints with different functionalities based on the cognitive characteristics of LLMs, which we refer to as the Hint-Library. Figure \ref{fig:hint-library} presents some representative hints from the Hint Library. Second, these hints are randomly inserted after certain high-frequency conjunctions, such as "Alternatively" and "Wait" (see more in Figure \ref{fig:Word cloud}), because these words typically indicate that the model begins to introspect or seek new solutions to the problem~\citep{search-o1}. Additionally, we also add hints before the stop token of long CoT, as this approach provides the LRM with more time to think without disrupting its original reasoning process. We find it intriguing that when hints are added before the stop token of Long Chain-of-thought (CoT), the model exhibits a sequential test time scaling effect; that is, as the thinking time increases, the success rate of problem-solving also gets higher(see more in \ref{hint-infer}). Through a series of data scoring, filtering, modifications, and rejection sampling fine-tuning (RFT)~\citep{rft,raft,star}, we eventually obtain our START from QwQ-32B-Preview. From Figure \ref{fig:maincase}, it can be seen that there is a comparison between the reasoning of QwQ-32B-Preview and START. When encountering a complex case analysis, QwQ-32-Preview generates hallucinations and provides incorrect answers, while START utilizes a code interpreter to self-debug, delivering the correct answer.

% Our approach is twofold: data synthesis and model fine-tuning. Initially, we introduce Hint-Reject Sampling Fine-Tuning (Hint-RFT), where actionable hints are interspersed within a subset of the training data to trigger tool usage during reasoning. This phase generates a dataset comprising instances where tool assistance transitions QwQ from an inability to solve a problem to achieving a successful resolution. Subsequently, through iterative Reject Sampling Fine-Tuning (RFT), we expand this dataset by deploying an intermediate model, START-Instruct, which further refines the tool-invocation capabilities across diverse problem scenarios, including multi-tool engagements.

% Moreover, to optimize the model's decision-making during tool-assisted reasoning, we develop START-RM, an outcome-based reward model trained on both successful and unsuccessful tool-assisted reasoning attempts. This reward model facilitates the selection of high-confidence responses in ambiguous situations and enhances the best-of-N performance during inference. Leveraging reinforcement learning techniques, we further fine-tune START-Instruct to produce START-RL, a refined version that exemplifies superior reasoning efficacy.

Empirical evaluations across a suite of benchmarks encompassing mathematical problem-solving, scientific inquiries, coding challenges, and GPQA tasks demonstrate that START-RL markedly surpasses existing tool-integrated and long CoT models, including QwQ-32B-Preview, o1-mini, and o1-preview in MATH benchmarks. These results underscore the efficacy of integrating external tools into the long CoT framework, highlighting START-RL as the first open-source tool-integrated long CoT reasoning model that sets a new standard for LLM performance in complex reasoning domains.

In summary, our contributions are threefold:
\begin{itemize}
    \item We introduce Hint-infer, a simple and effective sequential test-time scaling method for LRMs such as QwQ-32B-Preview and START.
    \item We introduce Hint-RFT, a self-training framework that enables a large language model (LRM) to teach itself how to utilize code interpreter.
    \item We present START, the first open-source LRM that utilizes long CoT and code interpreter to address complex reasoning tasks.
\end{itemize}

% START not only advances the state-of-the-art in tool-integrated reasoning, but also provides new insights into mitigating model hallucinations and enhancing reasoning efficiency, paving the way for more reliable and capable LLMs in diverse application domains.
\section{Related Work}
\label{sec:related}

Large Language Models have demonstrated remarkable dominance across numerous Natural Language Processing tasks.
To enhance the complex reasoning capabilities of LLMs, \citet{CoT} introduce Chain-of-Thought (CoT), which incorporates multi-step intermediate reasoning before arriving at final conclusions.
CoT exhibits significant advantages across multiple domains, including mathematics, science, and programming.
Subsequently, \citet{o1} further explore CoT and propose the Long Chain-of-Thought framework.
In Long CoT, LLMs demonstrate advanced cognitive behaviors such as reflection, verification, correction, and multi-path exploration, thereby further enhancing their problem-solving capabilities in complex reasoning tasks.
Moreover, Long CoT exhibits excellent test-time scaling properties, where increased computational resources correlate with improved reasoning outcomes.
Models like QwQ-32B-Preview~\citep{QwQ}, DeepSeek-R1~\citep{DeepSeekR1}, k1.5~\citep{k1.5},and InternThinker~\citep{cai2024internlm2} have successfully experimented with Long CoT for enhanced reasoning, combining fine-tuning and Reinforcement Learning to elevate the performance of open-source reasoning models to unprecedented levels.
Notably, subsequent models such as Open-R1~\citep{OpenR1}, O1Replication~\citep{O1Replication}, S1~\citep{s1simpletesttimescaling} and LIMO~\citep{limo} observes significant benefits from Long CoT even in smaller models through simple distillation.
\par
Nevertheless, significant challenges persist, particularly in addressing hallucination phenomena and computational inaccuracies that impede optimal performance.
Drawing parallels with human cognition, where external aids such as scratch paper and calculators substantially mitigate computational errors, LLMs can similarly benefit from the integration of auxiliary tools. 
Research by~\citet{deepseekmath} demonstrates that code-based pre-training protocols significantly augment LLMs' mathematical reasoning proficiency.
Various works successfully implemente Python-based computational tools to enhance model performance~\citep{pot,tora,MARIO,DotaMath}.
In the domain of mathematical proof verification, the incorporation of Lean yield notable advancements~\citep{DeepSeekProver,InternLM2p5StepProver}.
\par
This study synthesizes the advantages of Python-based tools and long CoT methodologies, advancing LRMs like QwQ-32B-Preview through the integration of tool utilization capabilities.
This integrated approach yields improved performance metrics across mathematical and coding benchmarks.
\section{Methodology}
\label{sec:method}

% \subsection{Overview}

% In this section, we present the methodologies employed in our study to enhance the performance of large models in complex reasoning tasks, particularly in the context of tool invocation and long-context learning. Previous work has demonstrated the efficacy Long CoT in tackling intricate reasoning problems; however, the specific use of tool invocation, an approach typically used to augment the capabilities of large models and mitigate hallucination has not been thoroughly explored within the Long CoT framework. A significant challenge lies in the synthesis of appropriate data for this application. In response, we introduce an innovative yet straightforward method called \textbf{Hint-RFT}, which involves the strategic insertion of hints during the model's reasoning process to activate its potential for tool invocation. Moreover, we leverage existing datasets generated during iterative Hint-RFT, specifically targeting Long CoT and TIR data, to train a Generative Reward Model (START-RM) that enhances the model's scaling capabilities during testing. Lastly, we apply a \textbf{reinforcement learning} algorithm, specifically focusing on optimizing the answer space through exploration, to further elevate the model's performance.

\subsection{Training data}
Our training data comprises two parts: one consists of math data sourced from previous AIME problems~\footnote{\url{https://huggingface.co/datasets/gneubig/aime-1983-2024}}(before 2024), MATH~\citep{math}, and Numina-MATH~\citep{numinamath}, while the other includes code data from Codeforces~\footnote{\url{https://codeforces.com/problemset}}, code contests~\footnote{\url{https://github.com/google-deepmind/code_contests}} and LiveCodeBench(before July 2024)~\citep{livecodebench}. We apply the same decontamination method as described in \citep{qwen2.5math} to the training set in order to minimize potential test data leakage risks. There are a total of 40K math problems and 10K code problems, and the specific quantity distribution can be referred to in Appendix \ref{sec: training data}.

\subsection{Hint-RFT}
\paragraph{Construct Hint}
we have designed a series of hints(Hint-Library) tailored to the various scenarios that may arise during LLM reasoning. Since mathematical reasoning with tools can be quite complex, we develop different hints focused on reflection, logical verification, exploring new methods, and more. These diverse hints enable the model to adopt different strategies based on the specific situation it encounters.
For coding tasks, we concentrate on designing hints that promote the model's self-debugging capabilities. By encouraging the model to check its code against test cases, it can verify the correctness of its solutions and make necessary adjustments as needed. We find that adding code template to the hint can effectively prompt the model to generate desired debugging code. Figure \ref{fig:hint-library} show the hints.

\paragraph{Hint-infer} 

For mathematical reasoning, we strategically insert hints after specific conjunction tokens, such as \textit{Alternatively} and \textit{Wait} as these tokens typically indicate that the model may be questioning its own reasoning or considering alternative approaches. It is important to note that after the hints are inserted, the model continues its reasoning process. The generated code is then sent to a Python interpreter for execution, and upon obtaining the results, the model proceeds to generate further outputs based on that information. Similarly, we can insert hints before the stop token to encourage the model to engage in deeper reasoning based on its existing reasoning. By inserting hints at these critical junctures at random, we encourage the model to explore a broader reasoning space. For code reasoning, we primarily concentrate on code generation tasks. We insert hints that prompt the model to test its own code right before the model generates the final code solution. This strategic placement encourages the model to engage in self-assessment, thereby enhancing the accuracy and reliability of the generated code. A more intuitive description is in Figure \ref{fig:mainfig}.

\paragraph{Data process and model fine-tuning}
Inspired by ~\citep{prm}, we adopt an active learning method, where we perform greedy inference and hint inference using QwQ-32B-Preview on all training data, and we recall data from reasoning tasks where QwQ-32B-Preview would not succeed without tools, but succeeded with hint inference. This is incorporated into our startup data $D_{seed}$ with 10K math data and 2K code data. It is important to note that, in addition to scoring the generated reasoning trajectories based on the rules, we also filter out responses that contain repetitive patterns. Additionally, we modify the Python identifiers in the code data hints to \textit{"Debug Code Template"} and remove the output placeholders.  We fine-tune QwQ-32B-Preview based on $D_{seed}$  to obtain START-0. The purpose of this fine-tuning step is to enable the model to learn the response paradigm for utilizing tools.

% Given that we have designed a diverse set of hints for mathematical reasoning, we categorize them based on their frequency of use. Each category is assigned a different insertion probability, with more common hints—such as those promoting reflection and self-checking—allocated a higher likelihood of occurrence.

\subsection{RFT}

To further enhance the diversity and quantity of the training data, as illustrated in Figure \ref{fig:mainfig}, we utilize the obtained START-0 to perform rejection sampling fine-tuning on all training data. Specifically, we use sampling parameters of temperature 0.6 and top-p 0.95 with START-0 to perform 16 rounds of sampling. We score the sampled long TIR data, filter out responses with repetitive patterns, and manually modify any unreasonable content. We retain a maximum of one response per question, resulting in our dataset $D_{\text{START}}$. Using the 40,000 math data entries and 10,000 code data entries from $D_{\text{START}}$, we fine-tune QwQ-32B-Preview once again, resulting in our final LRM named START.

\section{Experiment}
\label{sec:experiment}
\begin{table*}[!t]
\centering
\caption{Main results on challenging reasoning tasks, including PhD-level science QA, math, and code benchmarks. We report Pass@1 metric for all tasks. For models with 32B parameters, the best results are in \textbf{bold} and the second-best are \underline{underlined}. Symbol ``$^\dagger$'' indicates results from their official releases.}
\label{tab:reasoning_performance}
\setlength\tabcolsep{4pt} % Adjusted for fewer columns
\fontsize{8.1pt}{10.5pt}\selectfont
\begin{tabular}{p{2.65cm}cccccc}
\toprule
\textbf{Method} & \textbf{GPQA} & \textbf{MATH500} & \textbf{AMC23} & \textbf{AIME24} & \textbf{AIME25} & \textbf{LiveCodeBench} \\
\midrule
\multicolumn{7}{c}{\textit{\textbf{General LLMs}}} \\
\midrule
Qwen2.5-32B & 46.4 & 75.8 & 57.5 & 23.3 &- & 22.3 \\
Qwen2.5-Coder-32B & 33.8 & 71.2 & 67.5 & 20.0 &- & 25.0 \\
% Qwen2.5-MATH-72B & - & 88.1 & - & - & & - \\
% Qwen2.5-72B & {49.0} & {79.4} & {67.5} & {20.0} & & {33.0} \\
Llama3.3-70B & {43.4} & {70.8} & {47.5} & {36.7} & -& {34.8} \\
DeepSeek-V3-671B & {59.1} & {90.2} & {-} & {39.2} &- & {40.5} \\
GPT-4o$^\dagger$ & {50.6} & {60.3} & {-} & {9.3} &- & {33.4} \\
\midrule
\multicolumn{7}{c}{\textit{\textbf{Reasoning LLMs}}} \\
\midrule
\multicolumn{7}{l}{\textit{API Only}} \\
\midrule
o1-preview$^\dagger$ & {73.3} & {85.5} & {81.8} & {44.6} & 37.5 & {53.6} \\
o1-mini$^\dagger$ & {-} & {90.0} & {-} & {63.6} & 50.8 & {-} \\
o1$^\dagger$ & {77.3} & {94.8} & {-} & {74.4} & - & {63.4} \\
o3-mini(low)$^\dagger$ & {70.6} & {95.8} & {-} & {60.0} & 44.2 & {75.6} \\
% o3-mini(medium)$^\dagger$ & {76.8} & {97.3} & {-} & {79.6} & 66.7 & {84.6} \\
% o3-mini(high)$^\dagger$ & {79.7} & {97.9} & {-} & {87.3} & 76.7 & {82.0} \\
% DeepSeek-R1-Lite$^\dagger$ & {58.5} & {91.6} & {-} & {52.5} & - & {51.6} \\

\midrule
\multicolumn{7}{l}{\textit{Open weights}} \\
\midrule
% DeepSeek-R1$^\dagger$ & {58.5} & {91.6} & {-} & {52.5} & - & {51.6} \\
R1-Distill-Qwen-32B$^\dagger$ & {62.1} & {\underline{94.3}} & {\underline{93.8}} & {\textbf{72.6}} &\underline{46.7} & {\textbf{57.2}} \\
s1-32B$^\dagger$ & {59.6} & {93.0} & {-} & {50.0} & 33.3 & {-} \\
Search-o1-32B$^\dagger$ & {\textbf{63.6}} & {86.4} & {85.0} & {56.7} & - & {33.0} \\
QwQ-32B-Preview & {58.1} & {90.6} & {80.0} & {50.0} & 40.0& {41.4} \\
\midrule
% START (Ours) & $\text{\underline{63.6}}{\textcolor{deepgreen}{\text{(+3.5)}}}$ & \textbf{94} & \textbf{95.0} & \underline{66.7} & \textbf{46.7} & \underline{46.4} \\
START
& 
$\text{\textbf{63.6}}{\textcolor{deepgreen}{\text{(+5.5)}}}$ 
& 
$\text{\textbf{94.4}}{\textcolor{deepgreen}{\text{(+3.8)}}}$ 
& 
$\text{{\textbf{95.0}}}{\textcolor{deepgreen}{(+15.0)}}$ 
&
$\text{{\underline{66.7}}}{\textcolor{deepgreen}{(+16.7)}}$  
& 
$\text{{\textbf{47.1}}}{\textcolor{deepgreen}{\text{(+7.1)}}}$ 
& 
$\text{\underline{47.3}}{\textcolor{deepgreen}{\text{(+5.9)}}}$  \\
% START-Hint-infer
% & 
% $\text{\textbf{}}{\textcolor{deepgreen}{\text{(+3.5)}}}$ 
% & 
% $\text{\textbf{}}{\textcolor{deepgreen}{\text{(+3.4)}}}$ 
% & 
% $\text{{\textbf{}}}{\textcolor{deepgreen}{(+15.0)}}$ 
% &
% $\text{{\textbf{}}}{\textcolor{deepgreen}{(+16.7)}}$  
% & 
% $\text{{\textbf{60.0}}}{\textcolor{deepgreen}{\text{(+20.0)}}}$ 
% & 
% $\text{\underline{}}{\textcolor{deepgreen}{\text{()}}}$  \\
\bottomrule
\end{tabular}
\end{table*}

\subsection{Benchmarks}
In this work, we primarily focus on integrating Python tools into long CoT reasoning.
Given Python's effectiveness in enhancing computational and programming aspects of reasoning, we select several representative and challenging reasoning benchmarks to validate our methodology.
% These include:
\par
\textbf{GPQA:} This benchmark comprises 448 graduate-level multiple-choice questions authored by experts in biology, physics, and chemistry~\citep{gpqa}.
These questions present significant challenges, as even domain experts achieved less than 75\% accuracy in testing~\citep{o1}.
% We select this benchmark to comprehensively and authoritatively demonstrate our method's advantages in solving scientific problems.
% Compared to simpler problems, these challenging questions better align with our pursuit of AI development that aims to comprehensively surpass human capabilities.
\par
\textbf{Math Benchmarks:} Mathematical performance of LLMs remains a focal point for researchers.
In the mathematical domain, we select MATH500~\citep{prm} at the high school level, along with competition-level AMC23~\footnote{\url{https://huggingface.co/datasets/AI-MO/aimo-validation-amc}}, AIME24~\footnote{\url{https://huggingface.co/datasets/AI-MO/aimo-validation-aime}} and AIME25~\footnote{\url{https://huggingface.co/datasets/TIGER-Lab/AIME25}} as our evaluation datasets.
These datasets encompass various mathematical question types, including algebra, calculus, number theory, probability, and geometry, enabling a comprehensive assessment of LLMs' mathematical problem-solving capabilities.
\par
\textbf{LiveCodeBench:} This benchmark evaluates LLMs' programming capabilities, with test cases categorized into easy, medium, and difficult levels~\citep{livecodebench}. We choose 112 problems from August 2024 to November 2024 as the code benchmark. These questions are categorized as hard, medium, and easy based on difficulty.

% This testing framework effectively demonstrates START's advantages in coding, debugging, and troubleshooting.
\par
% In our selection of testing benchmarks, we consider both difficulty levels and maintained consistency with QwQ, thereby thoroughly demonstrating the performance improvements achieved by incorporating Python tools into long CoT reasoning for challenging inference tasks in large language models.
%

\subsection{Baselines}
We evaluate our approach against the following baseline methods:

\textbf{General LLMs:} These methods are general LLMs without the long CoT reasoning. The open-source models include Qwen2.5-32B-Instruct~\cite{qwen2.5}, Qwen2.5-Coder-32B-Instruct~\cite{qwen2.5coder},DeepSeek-V3-671B~\cite{deepseekv3}, Llama3.3-70B-Instruct~\cite{llama3} and GPT-4o~\citep{gpt4o}.

\textbf{LRMs:} These methods are equipped with long CoT reasoning. 
\textbf{(1) API only:} These models can only be accessed through the API, including o1-series~\citep{o1} and o3-mini~\citep{o3}.
\textbf{(2) Open weights:} we compare with some open weights LLMs, including DeepSeek-r1 series~\citep{DeepSeekR1}, QwQ-32B-Preview~\citep{QwQ}, s1~\citep{s1} and Search-o1~\citep{search-o1}.

% \subsection{Training Details}

\subsection{Implementation Details}

We fine-tune QwQ-32B-Preview with $D_\textbf{START}$ to get START. We train the base models with key settings including a 7e-6 learning rate, 128 global batch size, a cosine scheduler with a 3\% warm-up, a maximum context length of 16,384 tokens and 3 training epochs. Checkpoints are not selected with early stops.  The training process employs full-parameter fine-tuning with DeepSpeed ZeRO-3~\citep{zero} optimization.
Responses are generated using greedy decoding with a maximum sequence length of 32,768 and a limit of 6 maximum tool uses. For all benchmarks, we report the pass@1 performance and the evaluation metric is from ~\citep{qwen2.5math}. We use the same training and inference chat template as QwQ-32B-Preview~\citep{QwQ}.
% For MATH500,GPQA and LiveCodeBench, we report the greedy accuracy. For AMC23, AIME24 and AIME25, we report the average accuracy with temperature 0.5, top-p 0.95 and sample time 16. 
% We will explain why we do not use R1-Distill-Qwen-32B as base model or data generation model, si
% All models are trained with \textit{DeepSpeed ZeRO Stage3}\citep{deepspeed} and \textit{Flash-Attention 2}\citep{flashattention}. 
The hardware setup involves 32 NVIDIA A100 GPUs.

% \begin{table}[ht]
%     \centering
%     \caption{Scores on GPQA in various subjects.}
%     \begin{tabular}{@{}lccc@{}}
%         \toprule
%         Model      & Physics & Chemistry & Biology \\ \midrule
%         QwQ-32B-Preview   & 73.8    & 41.9      & 68.4    \\
%         Search-o1 & 77.9    & \textbf{47.3}      & \textbf{78.9}    \\
%         START     & \textbf{80.0}    & \textbf{47.3}      & 68.4    \\ \bottomrule
%     \end{tabular}
%     \label{tab:gpqa_scores}
% \end{table}
\begin{table}[ht]
    \centering
    \caption{Scores on GPQA in various subjects.}
    \setlength{\tabcolsep}{2pt} % 控制列间距
    \begin{tabular}{lccc}
        \toprule
        Model      & Physics & Chemistry & Biology \\ \midrule
        QwQ-32B-Preview & 73.8 & 41.9 & 68.4 \\
        Search-o1 & 77.9 & \textbf{47.3} & \textbf{78.9} \\
        START     & \textbf{80.0} & \textbf{47.3} & 68.4 \\ \bottomrule
    \end{tabular}
    \label{tab:gpqa_scores}
\end{table}

\begin{table}[ht]
    \centering
    \caption{Scores on questions of different difficulty levels on LiveCodeBench.}
    \begin{tabular}{@{}lccccc@{}}
        \toprule
        Model      & Easy & Medium & Hard \\ \midrule
        QwQ-32B-Preview   & \textbf{92.3}    & 46.0      & 10.2    \\
        START     & \textbf{92.3}    & \textbf{84.6}     & \textbf{12.2}    \\ \bottomrule
    \end{tabular}
    \label{tab:lcb_scores}
\end{table}

\begin{table*}[!t]
\centering
\caption{Compare long cot with long tir on challenging reasoning tasks, including PhD-level science QA, math, and code benchmarks. We report Pass@1 metric for all tasks.}
\label{tab:longtirvscot}
\setlength\tabcolsep{4pt} % Adjusted for fewer columns
\fontsize{8.1pt}{10.5pt}\selectfont
\begin{tabular}{p{2.65cm}cccccc}
\toprule
\textbf{Method} & \textbf{GPQA} & \textbf{MATH500} & \textbf{AMC23} & \textbf{AIME24} & \textbf{AIME25} & \textbf{LiveCodeBench} \\
% \midrule
\midrule
QwQ-32B-Preview & {58.1} & {90.6} & {80.0} & {50.0} & 40.0& {41.4} \\
QwQ-RFT & {58.5} & {{91.8}} & {{82.5}} & {{53.3}} &{33.3} & {{42.1}} \\
\midrule
% START (Ours) & $\text{\underline{63.6}}{\textcolor{deepgreen}{\text{(+3.5)}}}$ & \textbf{94.4} & \textbf{95.0} & \underline{66.7} & \textbf{47.1} & \underline{46.4} \\
START (Ours) 
& 
$\text{\textbf{63.6}}{\textcolor{deepgreen}{\text{(+5.5)}}}$ 
& 
$\text{\textbf{94.4}}{\textcolor{deepgreen}{\text{(+3.8)}}}$ 
& 
$\text{{\textbf{95.0}}}{\textcolor{deepgreen}{(+15.0)}}$ 
&
$\text{{\textbf{66.7}}}{\textcolor{deepgreen}{(+16.7)}}$  
& 
$\text{{\textbf{47.1}}}{\textcolor{deepgreen}{\text{(+7.1)}}}$ 
& 
$\text{{\textbf{47.3}}}{\textcolor{deepgreen}{\text{(+5.9)}}}$  \\
\bottomrule
\end{tabular}
\end{table*}
\subsection{Main Results}

Table~\ref{tab:reasoning_performance} presents the evaluation results of START across various benchmarks, demonstrating its superior reasoning performance in scientific, mathematical, and coding domains compared to other open-source models. Overall, general LLMs, even domain-specific LLMs, are difficult to compete with LRMs in complex tasks.

\paragraph{PHD-level Science QA Performance}
It can be observed that on the ScienceQA benchmark, START demonstrates an absolute improvement of 5.5\% over QwQ-32B-Preview, achieving the same score as the state-of-the-art model, search-o1-32B. Table \ref{tab:gpqa_scores} presents the scores of QwQ-32B-Preview, Search-o1, and START across three subjects of GPQA: Physics, Chemistry, and Biology. Specifically, START achieves the highest score in Physics, while Search-o1 outperforms QwQ-32B-Preview significantly in Biology. This discrepancy can be attributed to the fact that Physics often necessitate extensive computational reasoning, whereas Biology primarily relies on knowledge-based reasoning. Consequently, the utilization of Python-based tools(START) yields more pronounced efficacy in the former disciplines, while the utilization of internet knowledge(search-o1-32B) works better on the latter. 

\paragraph{MATH Benchmarks Performance}
On the MATH benchmarks, START also demonstrates considerable advantages over QwQ-32B-Preview. Specifically, it achieves absolute improvements of 3.8\%, 15.0\%, 16.7\% and 7.1\% on the MATH500, AMC23, AIME24 and AIME25, respectively. The performance of START is comparable to that of R1-Distill-Qwen-32B, which is distilled from 671B DeepSeek-R1, and overall it exceeds o1-preview. These results highlight the significant role of Python-based tools in enhancing mathematical reasoning capabilities.

\paragraph{LiveCodeBench Performace}
On the LiveCodeBench, START, by equipping the model with the capability to invoke debugging tools, achieves an absolute improvement of 5.9\% over QwQ-32B-Preview. We find that START improves the most compared to QwQ-32B-Preview on questions with medium difficulty from \ref{tab:lcb_scores}.
The possible reason is that for easy questions, QwQ-32B-Preview can generate the correct answers with high probability without debugging, and for hard questions, based on the current capabilities of the model, a limited number of debugs is also difficult to solve.
% and it also outperforms search-o1-32B and DeepSeek-V3-671B.

% This phenomenon can be attributed to the fact that, for simpler problems, the probability of generating correct code in a single attempt is higher, thereby diminishing the relative necessity for debugging. As a result, START's efficacy becomes increasingly evident as the complexity of the tasks escalates.

\subsection{Analysis}

\subsubsection{Long CoT vs Long TIR}

\begin{figure*}[ht]
    \centering
    \includegraphics[width=1.0\linewidth]{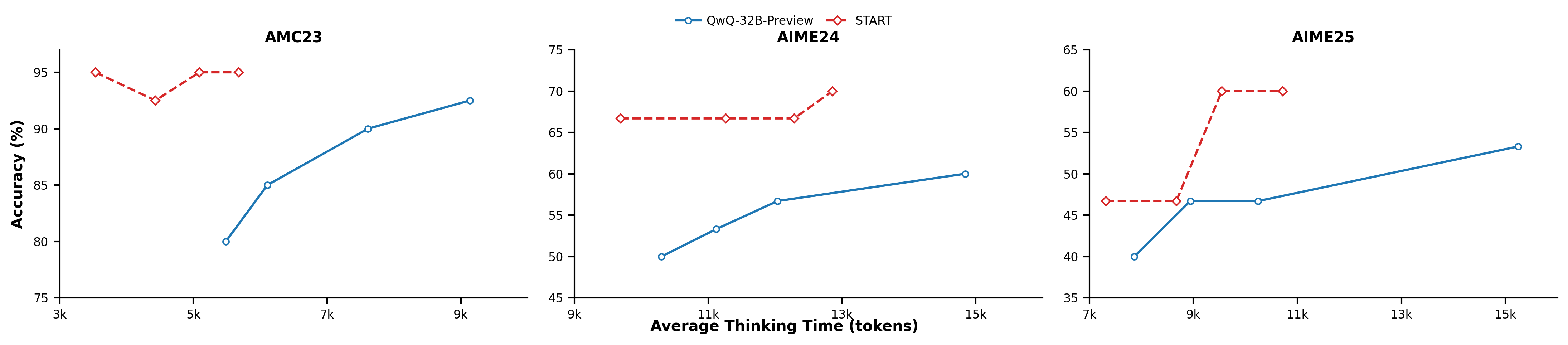}
    \caption{Test time scaling for QwQ-32B-Preview and START on challenge math bench marks via Hint-infer.}
    \label{fig:testtimescaling}
\end{figure*}

To ascertain whether our performance gains stem from the additional training questions or from the tool invocation capability, we conduct an experiment using the same set of queries from $D_{START}$ but only apply RFT with QwQ-32B-Preview. For each query, we sampled 32 responses with a temperature of 0.7 and a top-p value of 0.95. After filtering out incorrect responses and  responses with repeating strings, we retain at most one response per question and get the long CoT dataset $D_{\text{RFT}}$. Based on $D_{\text{RFT}}$, we fine-tune QwQ-32B-Preview, yielding QwQ-RFT. This methodological approach allows us to isolate the impact of tool invocation from that of the expanded training dataset.

The results presented in Table~\ref{tab:longtirvscot} indicate that the performance of QwQ-RFT is nearly on par with that of QwQ-32B-Preview. Therefore, the observed performance advantage of START is likely predominantly driven by its tool invocation capability, suggesting that this feature plays a critical role in enhancing its effectiveness.

\subsubsection{Analysis of Hint-infer}
\label{hint-infer}

\paragraph{Compare QwQ-32B-Preview with Hint-infer and START}
Through Hint-RFT, we discover that QwQ-32B-Preview inherently possesses the potential to invoke tools, although this capability is challenging to elicit through prompting alone and instead requires explicit hints to activate. START, which is fine-tuned from QwQ-32B-Preview using Hint-RFT, allows us to directly compare the performance of QwQ-32B-Preview with Hint-infer against that of START. To avoid interrupting QwQ-32B-Preview's reasoning process, we only insert hints before the stop token of QwQ-32B-Preview(see more in Appendix \ref{hint-infer-prompt}).  This comparison provides a basis for evaluating the necessity of fine-tuning, as it helps to determine whether the enhanced performance of START is primarily due to the fine-tuning process or can be sufficiently achieved through hint-based prompting alone.

From Table ~\ref{tab:hint-all-results}, it is evident that incorporating hints during the inference process of QwQ-32B-Preview leads to improvements across all benchmarks. However, these improvements are relatively modest compared to the gains achieved by START. Consequently, START, through Hint-RFT, significantly enhances QwQ-32B-Preview's tool invocation capabilities, demonstrating the effectiveness of fine-tuning in unlocking  the model's latent potential.

\paragraph{Test-time scaling via Hint}

By inserting hints at the end of QwQ-32B-Preview's inference process, we can simultaneously increase both the model's thinking time and its accuracy(see Figure \ref{fig:testtimescaling}) by multiple rounds of inserting hint before stop token. It indicates that Hint-infer is a simple yet effective method for achieving sequential test-time scaling. Unlike the strategy outlined in ~\citep{s1}, which merely increases the number of "wait" tokens, our method augments the number of tool invocation. The use of Hint-infer for START, on the other hand, does not work as well as on QwQ-32B-Preview, and the reason behind this may be that those hints we inserted were already available during the reasoning process between STARTs, reducing the amount of information in the added hints. More results and analysis are list in \ref{hint-infer-more}.

\section{Conclusion}
\label{sec: conclusion}

this paper presents START, a groundbreaking tool-integrated long Chain-of-Thought reasoning model that effectively mitigates the limitations of existing large reasoning models (LRMs) through the innovative integration of external tools and self-learning techniques. Our contributions, namely Hint-infer and Hint-RFT, showcase a novel approach to enhance reasoning capabilities by enabling LRM to leverage coding interpreters for complex computations and self-debugging. The empirical results demonstrate significant improvements in performance across a range of challenging benchmarks, establishing START as a leading open-source solution for advanced reasoning tasks. By combining long CoT with tool integration, START sets a new standard for the future development of LLMs, paving the way for more reliable and efficient reasoning in higher-level cognitive tasks.

\newpage
\section{Limitations}
While our work on START demonstrates significant advancements in tool-integrated long Chain-of-Thought reasoning, it is essential to acknowledge several limitations inherent in our approach.

Firstly, our research exclusively focuses on the integration of a Python interpreter as the sole external tool. Although this choice was made for its relevance to many reasoning tasks, we believe that incorporating a wider variety of tools—such as search engines, specialized libraries, or different computational resources—could potentially enhance the model's performance and versatility. Future work could explore how diverse toolsets might contribute to more robust reasoning across various domains.

Secondly, the manual design of hints for insertion into the long CoT reasoning process may inadvertently disrupt the model’s original flow of thought. While we aimed to strategically position these hints to optimize performance, the effectiveness of hint positioning and selection could vary based on the specific task or context. More nuanced criteria for determining the most effective types and placement of hints might result in further improvements in reasoning fluidity and accuracy.

Additionally, our empirical evaluations were conducted on a limited set of benchmarks. Although results reported demonstrate promising outcomes, the generalizability of our findings remains to be established across broader and more diverse datasets. The performance of START may be sensitive to variations in task complexity, domain specificity, and the characteristics of the input data.

% Moreover, while we have introduced mechanisms to improve reliability and reduce hallucinations, there is still a risk of model biases, particularly in the context of the tasks for which our model was evaluated. Addressing biases in model outputs and understanding their implications on different demographic groups remains an important area for further investigation.

Lastly, potential risks associated with the misuse of the technology must be considered. The ability of our model to generate code or suggest problem-solving strategies could be inadvertently leveraged for malicious purposes, such as crafting disinformation or automating harmful tasks. It is crucial to implement safeguards and establish ethical guidelines to monitor and mitigate such risks.

In summary, while our research provides a significant step forward, acknowledging these limitations is essential to paving the way for future improvements and ensuring the responsible development and application of tool-integrated reasoning models.

\bibliography{custom}

\newpage
\newpage
\appendix

% \section{Example Appendix}
\section{Appendix}

\subsection{Training set of START}
\label{sec: training data}

\begin{table*}[ht]
    \centering
    \caption{Sources of Dataset D}
    \begin{tabular}{lc}
        \toprule
        Source & Quantity \\
        \midrule
        AIME problems (before 2024) & 890 \\
        MATH~\citep{math} & 7500 \\
        Numina-MATH~\citep{numinamath} & 28505 \\
        \midrule
        \textbf{Code Data} & \\
        Codeforces & 7505 \\
        Code contests & 2011 \\
        LiveCodeBench (before July 2024)~\citep{livecodebench} & 558 \\
        \midrule
        \textbf{Total} & 49969 \\
        \bottomrule
    \end{tabular}
\end{table*}

\subsection{More results about Hint-infer}
\label{hint-infer-more}

Building upon the observed trends, the detailed results in Table \ref{hint-infer-more} further underscore the efficacy of the QwQ-Hint-infer and START-Hint-infer methods across diverse challenging reasoning tasks. Specifically, for datasets such as aime24, aime25, gpqa, amc23, MATH500, and LiveCodeBench, QwQ-32B-Preview consistently demonstrates performance enhancements with each subsequent round of hint insertion. For instance, aime24 improves from 50.0\% in Round 0 to 60.0\% in Round 3, and MATH500 shows a marginal yet steady increase from 90.6\% to 92.4\% over the same rounds. This consistent upward trend highlights the method's ability to incrementally refine the model's reasoning capabilities through iterative hint integration.

In contrast, the START-Hint-infer approach exhibits a more varied performance across different datasets. While there are improvements in some areas, such as AIME25, where the Pass@1 metric reaches 60.0\% by Round 3 and LiveCodeBench sees an increase from 47.3\% to 50.0\%, other datasets like GPQA and LiveCodeBench show relatively modest gains and even no gains. This disparity suggests that the effectiveness of Hint-infer may be contingent on the inherent characteristics of the dataset and the nature of the reasoning tasks involved.

\begin{table*}[ht]
    \centering
    \caption{Comparison of QWQ-Hint-infer and START-Hint-infer on challenging reasoning tasks, including PhD-level science QA, math, and code benchmarks. We report Pass@1 metric for all tasks.}
    \resizebox{0.8\textwidth}{!}{ 
    \begin{tabularx}{\textwidth}{@{}lXXXX|XXXX@{}}
        \toprule
        \textbf{Dataset} & \multicolumn{4}{c|}{\textbf{QwQ-32B-Preview}}                   & \multicolumn{4}{c}{\textbf{START}} \\ \cline{2-9}
                         & \textbf{Round 0} & \textbf{Round 1} & \textbf{Round 2} & \textbf{Round 3} & \textbf{Round 0} & \textbf{Round 1} & \textbf{Round 2} & \textbf{Round 3} \\ \midrule
        aime24          & 50.0\%          & 53.3\%          & 56.7\%          & 60.0\%          & 66.7\%          & 66.7\%          & 66.7\%          & 70.0\%          \\
        aime25          & 40.0\%          & 47.1\%          & 47.1\%          & 53.3\%          & 47.1\%          & 47.1\%          & 60.0\%          & 60.0\%          \\
        gpqa            & 58.5\%          & 58.6\%          & 59.6\%          & 59.6\%          & 63.6\%          & 61.6\%          & 60.6\%          & 61.6\%          \\
        amc23           & 80.0\%          & 85.0\%          & 90.0\%          & 92.5\%          & 95.0\%          & 92.5\%          & 95.0\%          & 95.0\%          \\
        MATH500         & 90.6\%          & 92.0\%          & 92.0\%          & 92.4\%          & 94.4\%          & 95.0\%          & 95.6\%          & 95.2\%          \\
        LiveCodeBench   & 41.4\%          & 42.0\%          & 42.0\%          & 42.0\%          & 47.3\%          & 48.2\%          & 50.0\%          & 50.0\%          \\ 
        \bottomrule
    \end{tabularx}
    }
    \label{tab:hint-all-results}
\end{table*}

% \section{Some experiments on R1-Distill-qwen-32B}
% \label{R1-32B-hint-infer}

% \section{Details about evaluation in Table \ref{tab:reasoning_performance}}
% \label{evaluation detail}

% \section{Data clean details}
% \label{dataclean}

\subsection{Prompting Methods for Data annotation}
\label{prompting method}

We investigated three common methods to trigger existing reasoning LLMs to generate long CoT with Python tool calls in mathematical reasoning tasks. The first method is "direct prompt"(\textit{Please integrate natural language reasoning with programs to solve the problem.}), which instructs the model to directly use Python tools during reasoning. The second method, "well-designed prompt" is derived from search-o1~\citep{search-o1} and provides detailed instructions on how to use the tools; this prompt successfully triggers the model to generate special tokens for browser calls in search-o1. The third method is "in-context prompt"(Give some demonstrations in the prompt) which leverages examples to guide the model in generating data in the same format.
We do not use general LLMs, as they typically cannot produce long CoTs. For the O1 series, we can only assess whether the summary includes Python tool invocation. As a result, we found that neither QwQ-32B-Preview, DeepSeek-R1, nor o1-mini could successfully generate long CoTs with code interpreter calls using the three prompt-based methods. In contrast, the hint-infer method was able to trigger the model to produce Python code with 100\% success.

\subsection{Hint-infer for test time scaling}
\label{hint-infer-prompt}

The three rounds hints of GPQA and MATH for Hint-infer are: \textit{Wait, I can use Python to check if my approach is correct and refine it, if necessary.```python}, \textit{Wait, I need to utilize Python code again to meticulously check to make sure I understand the question correctly as well as reasoning correctly.```python} and \textit{Wait, I can think more deeply about this problem through python tools.```python}. Hints of LivecodeBench is the same in Hint-Library. 
For code problem with starter code, the code template is 
\begin{verbatim}
{startcoder}
# Test the example inputs
solution = Solution()
# Example input1
test_input1 = ...
# Example input2
test_input2 = ...
# Print output
print(solution.function_name(test_input1))
print(solution.function_name(test_input2))  
# Check the output
'''
'''output
[...]
'''
\end{verbatim}
For code problem without starter code, the code template is 
\begin{verbatim}
def function_name(parameters):   
#Implementation\n
# Test the example inputs
solution = Solution()
# Example input1
test_input1 = ...
# Example input2
test_input2 = ...
# Print output
print(solution.function_name(test_input1))
print(solution.function_name(test_input2))  
# Check the output
'''
'''output
[...]
'''
\end{verbatim}

\newpage

\onecolumn
% \subsection{Case Study For START}

\subsection{Case Study For START}
\label{casestudy}

\subsubsection{A case form AMC23}
\begin{tcolorbox}[
colback=white,
colframe=black,
title=A case form AMC23,
breakable
]
\textcolor{red}{\textit{\textbf{Problem}:}}\\
In the state of Coinland, coins have values \(6\), \(10\), and \(15\) cents. Suppose \(x\) is the value in cents of the most expensive item in Coinland that cannot be purchased using these coins with exact change. What is the sum of the digits of \(x\)?

\textcolor{red}{\textit{\textbf{Solution}:}}\\
So I've got this problem from the state of Coinland, where they have coins worth \(6\), \(10\), and \(15\) cents. The question is asking for the value of the most expensive item that you can't buy using these coins with exact change. Then, I need to find the sum of the digits of that value.

First off, I need to understand what this is asking. It seems like a problem related to the concept of the Frobenius coin problem, which is about finding the largest amount that cannot be obtained using given coin denominations. In the classic case, with two coprime denominations, there's a formula to find this number, but here we have three denominations: \(6\), \(10\), and \(15\) cents.

I know that for two coprime numbers, \(a\) and \(b\), the largest number that can't be expressed as a non-negative integer combination of \(a\) and \(b\) is \(ab - a - b\). But here, we have three numbers, and they aren't all pairwise coprime. For instance, \(6\) and \(10\) share a common factor of \(2\), and \(6\) and \(15\) share a common factor of \(3\). So, the standard Frobenius number formula doesn't directly apply.

Maybe I should start by looking at the greatest common divisor (GCD) of these numbers. Let's find the GCD of \(6\), \(10\), and \(15\).

The GCD of \(6\) and \(10\) is \(2\).

The GCD of \(2\) and \(15\) is \(1\).

So, the overall GCD is \(1\). That means that eventually, any sufficiently large number can be expressed as a combination of these coins, since their GCD is \(1\).

But, there might be some gaps before we reach a certain point where all numbers can be expressed.

Perhaps I should look for the Frobenius number for these three denominations. However, I recall that for more than two denominations, there isn't a general formula for the Frobenius number. So, I might need to approach this problem differently.

One approach could be to consider the possible combinations of these coins and see which numbers can be expressed. Maybe I can start by listing out the possible sums.

Let's list out some combinations:

- \(0\) coins: \(0\) cents

- \(1\) coin: \(6\), \(10\), or \(15\) cents

- \(2\) coins: \(6 + 6 = 12\), \(6 + 10 = 16\), \(6 + 15 = 21\), \(10 + 10 = 20\), \(10 + 15 = 25\), \(15 + 15 = 30\) cents

- \(3\) coins: \(6 + 6 + 6 = 18\), \(6 + 6 + 10 = 22\), \(6 + 6 + 15 = 27\), \(6 + 10 + 10 = 26\), \(6 + 10 + 15 = 31\), \(6 + 15 + 15 = 36\), \(10 + 10 + 10 = 30\), \(10 + 10 + 15 = 35\), \(10 + 15 + 15 = 40\), \(15 + 15 + 15 = 45\) cents

- And so on...

This could take a while, and I might miss some numbers. Maybe there's a better way.

Another thought: since the GCD is \(1\), I know that every number greater than or equal to a certain number can be expressed as a combination of these coins. I need to find the largest number that cannot be expressed this way.

Perhaps I can look at the problem in terms of modulo arithmetic. For example, since \(6\) is divisible by \(3\), \(10\) is not, and \(15\) is divisible by \(3\), I can consider how these combine in terms of modulo \(3\).

Wait, but that might not be directly helpful. Let me think differently.

I've heard of the concept of the "Frobenius number" for two coins, but for three or more, it's more complicated. Maybe I can reduce this problem to a combination of smaller problems.

Let me consider pairs of these denominations and find their Frobenius numbers, then see how that helps.

First, consider only \(6\) and \(10\) cent coins. Since \(\text{GCD}(6,10)=2\), which is greater than \(1\), the Frobenius number for these two is not defined because you can't make all even numbers greater than a certain point with just \(6\) and \(10\). Wait, no. Actually, since their GCD is \(2\), you can make any even number greater than or equal to a certain value.

Wait, the Frobenius number formula \(ab - a - b\) applies only when \(a\) and \(b\) are coprime. Here, \(6\) and \(10\) are not coprime, so that formula doesn't apply.

Similarly, for \(6\) and \(15\), \(\text{GCD}(6,15)=3\), which is greater than \(1\).

And for \(10\) and \(15\), \(\text{GCD}(10,15)=5\).

So, none of the pairs are coprime.

This complicates things.

Alternatively, maybe I can look at the problem in terms of linear combinations.

I need to find the largest \(x\) such that there are no non-negative integers \(a\), \(b\), \(c\) satisfying \(6a + 10b + 15c = x\).

This seems tricky. Maybe I can look for a pattern in the possible sums.

Let me try to list out the possible sums in order:

\(0, 6, 10, 12, 15, 16, 18, 20, 21, 22, 24, 25, 26, 27, 30,\) etc.

Wait, I can use Python to generate specific examples that fit the problem's conditions.

\begin{tcolorbox}[colback=cyan!5!white, colframe=cyan!75!black, title=Python Code]

\begin{verbatim}
from itertools import product
def can_be_paid(x, coins):
    for a in range(x//coins[0] + 1):
        for b in range(x//coins[1] + 1):
            for c in range(x//coins[2] + 1):
                if a*coins[0] + b*coins[1] + c*coins[2] == x:
                    return True
    return False
coins = [6, 10, 15]
max_x = 100  
# arbitrary upper limit to search for the largest x that cannot be paid
possible_x = []
for x in range(1, max_x):
    if not can_be_paid(x, coins):
        possible_x.append(x)

print((max(possible_x), sum(int(digit) for digit in str(max(possible_x)))))
\end{verbatim}
\end{tcolorbox}

\begin{tcolorbox}[colback=orange!5!white, colframe=orange!75!black, title=Output]
\begin{verbatim}
(29, 11)
\end{verbatim}
\end{tcolorbox}

From the code, I found that the largest value that cannot be purchased using coins of \(6\), \(10\), and \(15\) cents is \(29\) cents. The sum of the digits of \(29\) is \(2 + 9 = 11\).

So, the answer is \(11\).

**Final Answer**
\[
\boxed{11}
\]

\end{tcolorbox}

\newpage
\subsubsection{A case form AIME24}
\begin{tcolorbox}[
colback=white,
colframe=black,
title=A case form AIME24,
breakable
]
\textcolor{red}{\textit{\textbf{Problem}:}}\\
Let \( p \) be the least prime number for which there exists a positive integer \( n \) such that \( n^{4}+1 \) is divisible by \( p^{2} \). Find the least positive integer \( m \) such that \( m^{4}+1 \) is divisible by \( p^{2} \).

\textcolor{red}{\textit{\textbf{Solution}:}}\\

I'm trying to solve this problem. It says: Let \( p \) be the least prime number for which there exists a positive integer \( n \) such that \( n^4 + 1 \) is divisible by \( p^2 \). Find the least positive integer \( m \) such that \( m^4 + 1 \) is divisible by \( p^2 \).

First, I need to find the smallest prime \( p \) such that there's some positive integer \( n \) making \( n^4 + 1 \) divisible by \( p^2 \). Then, using that \( p \), I have to find the smallest positive integer \( m \) where \( m^4 + 1 \) is divisible by \( p^2 \).

So, to start, I need to understand when \( n^4 + 1 \) is divisible by \( p^2 \). That means \( p^2 \) divides \( n^4 + 1 \), or in other words, \( n^4 \equiv -1 \pmod{p^2} \).

I know that for primes, certain properties hold due to Fermat's Little Theorem and Euler's Theorem, but I'm not sure how to apply them here directly because we're dealing with \( p^2 \), not just \( p \).

Maybe I should look at the equation \( n^4 \equiv -1 \pmod{p^2} \). This implies that \( n^8 \equiv 1 \pmod{p^2} \), because if \( n^4 \equiv -1 \), then squaring both sides gives \( n^8 \equiv 1 \pmod{p^2} \).

So, the order of \( n \) modulo \( p^2 \) divides 8, meaning the order is 1, 2, 4, or 8. But since \( n^4 \equiv -1 \pmod{p^2} \), the order can't be 1, 2, or 4, because in those cases, \( n^4 \) would be 1, not -1. Therefore, the order must be 8.

This means that 8 divides the order of the multiplicative group modulo \( p^2 \), which is \( \phi(p^2) = p(p-1) \), where \( \phi \) is Euler's totient function.

So, 8 divides \( p(p-1) \). Since \( p \) is prime, there are a few cases to consider:

1. \( p = 2 \)

2. \( p \) is an odd prime, and 8 divides \( p-1 \)

Let's consider \( p = 2 \) first.

If \( p = 2 \), then \( p^2 = 4 \). We need to check if there exists an \( n \) such that \( n^4 + 1 \) is divisible by 4.

Let's compute \( n^4 + 1 \) modulo 4:

- If \( n \) is even, \( n^4 \) is divisible by 16, so \( n^4 \equiv 0 \pmod{4} \), thus \( n^4 + 1 \equiv 1 \pmod{4} \), which is not divisible by 4.

- If \( n \) is odd, \( n^4 \equiv 1 \pmod{4} \), so \( n^4 + 1 \equiv 2 \pmod{4} \), which is also not divisible by 4.

Therefore, there is no positive integer \( n \) such that \( n^4 + 1 \) is divisible by 4. So, \( p = 2 \) is not a valid choice.

Next, consider odd primes where 8 divides \( p-1 \). That is, \( p \equiv 1 \pmod{8} \).

The smallest such primes are 17, 41, 73, etc.

Alternatively, there could be other primes where \( p \) divides \( n^4 + 1 \), but for \( p^2 \) to divide \( n^4 + 1 \), it's more restrictive.

Wait a minute, maybe I should consider quadratic residues or higher power residues.

Alternatively, perhaps I should look for primes where \( -1 \) is a fourth power residue modulo \( p^2 \).

I recall that for primes, \( -1 \) is a quadratic residue if \( p \equiv 1 \pmod{4} \), but here we're dealing with fourth powers.

Maybe I should consider the Legendre symbol or higher power residues.

Alternatively, perhaps I can use Hensel's Lemma to lift solutions from modulo \( p \) to modulo \( p^2 \).

First, I need to find primes \( p \) such that there exists an \( n \) with \( n^4 \equiv -1 \pmod{p} \), and then check if that solution can be lifted to modulo \( p^2 \).

So, start by finding primes \( p \) for which \( n^4 \equiv -1 \pmod{p} \) has a solution.

This is equivalent to saying that \( -1 \) is a fourth power residue modulo \( p \).

The multiplicative group modulo \( p \) is cyclic of order \( p-1 \), so \( n^4 \equiv -1 \pmod{p} \) has a solution if and only if there exists an integer \( k \) such that \( 4k \equiv \frac{(p-1)}{2} \pmod{p-1} \), since \( -1 \) is a square modulo \( p \) if \( p \equiv 1 \pmod{4} \).

Wait, I can use Python to automate these extensive calculations.

\begin{tcolorbox}[colback=cyan!5!white, colframe=cyan!75!black, title=Python Code]
\begin{verbatim}
def find_least_p():
    p = 3  # Start with the smallest odd prime
    while True:
        if any(pow(n, 4, p) == p - 1 for n in range(1, p)):
            # Check if n^4 = -1 mod p has a solution
            for n in range(1, p):
                if (n**4) % p == p - 1:
                    # Now check if there exists m such that m^4 = -1 mod p^2
                    p_squared = p * p
                    for m in range(1, p_squared):
                        if (m**4) % p_squared == p_squared - 1:
                            return p, m
        p = next_prime(p)

def next_prime(p):
    # Implement a function to get the next prime after p
    if p == 2:
        return 3
    candidate = p + 2
    while not is_prime(candidate):
        candidate += 2
    return candidate

def is_prime(num):
    # Implement a function to check if a number is prime
    if num <= 1:
        return False
    if num == 2:
        return True
    if num % 2 == 0:
        return False
    for i in range(3, int(num**0.5) + 1, 2):
        if num % i == 0:
            return False
    return True

# Find the least prime p and the least m
p, m = find_least_p()
print((p, m))
\end{verbatim}
\end{tcolorbox}

\begin{tcolorbox}[colback=orange!5!white, colframe=orange!75!black, title=Output]
\begin{verbatim}
SyntaxError: invalid syntax
\end{verbatim}
\end{tcolorbox}

It seems there was a syntax error in the code due to the way the condition was written. Let's correct the code and try again.

I need to ensure that the condition for checking if \( n^4 \equiv -1 \pmod{p} \) is properly implemented. Also, I need to make sure that the function to find the next prime is correctly defined.

Let's fix the code accordingly.

\begin{tcolorbox}[colback=cyan!5!white, colframe=cyan!75!black, title=Python Code]
\begin{verbatim}
def find_least_p():
    p = 3  # Start with the smallest odd prime
    while True:
        # Check if there exists n such that n^4 = -1 mod p
        found_n = False
        for n in range(1, p):
            if (n**4) % p == p - 1:
                found_n = True
                break
        if found_n:
            # Now check if there exists m such that m^4 = -1 mod p^2
            p_squared = p * p
            for m in range(1, p_squared):
                if (m**4) % p_squared == p_squared - 1:
                    return p, m
        p = next_prime(p)

def next_prime(p):
    # Implement a function to get the next prime after p
    if p == 2:
        return 3
    candidate = p + 2
    while not is_prime(candidate):
        candidate += 2
    return candidate

\end{verbatim}
\end{tcolorbox}

\begin{tcolorbox}[colback=cyan!5!white, colframe=cyan!75!black, title=Python Code]
\begin{verbatim}

def is_prime(num):
    # Implement a function to check if a number is prime
    if num <= 1:
        return False
    if num == 2:
        return True
    if num % 2 == 0:
        return False
    for i in range(3, int(num**0.5) + 1, 2):
        if num % i == 0:
            return False
    return True

# Find the least prime p and the least m
p, m = find_least_p()
print((p, m))
\end{verbatim}
\end{tcolorbox}

\begin{tcolorbox}[colback=orange!5!white, colframe=orange!75!black, title=Output]
\begin{verbatim}
(17, 110)
\end{verbatim}
\end{tcolorbox}

The code has determined that the least prime \( p \) for which there exists a positive integer \( n \) such that \( n^4 + 1 \) is divisible by \( p^2 \) is \( p = 17 \), and the least positive integer \( m \) such that \( m^4 + 1 \) is divisible by \( 17^2 = 289 \) is \( m = 110 \).

To verify this, let's check if \( 110^4 + 1 \) is indeed divisible by 289.

First, compute \( 110^4 \):

\[
110^2 = 12100
\]

\[
110^4 = 12100^2 = 146,410,000
\]

Then, \( 110^4 + 1 = 146,410,001 \).

Now, divide by 289:

\[
146,410,001 \div 289
\]

Let's perform the division:

\[
289 \times 506,613 = 289 \times 500,000 = 144,500,000
\]

\[
289 \times 6,613 = 289 \times 6,000 = 1,734,000
\]

\[
289 \times 613 = 289 \times 600 = 173,400
\]

\[
289 \times 13 = 3,757
\]

Adding them up:

\[
144,500,000 + 1,734,000 = 146,234,000
\]

\[
146,234,000 + 173,400 = 146,407,400
\]

\[
146,407,400 + 3,757 = 146,411,157
\]

Wait, that's higher than 146,410,001. Maybe I miscalculated the multiplication.

Alternatively, perhaps I should compute \( 146,410,001 - 146,411,157 = -1,156 \), which is not zero, indicating that 289 does not divide 146,410,001.

This suggests that there might be an error in the code's result.

Alternatively, maybe I need to double-check the calculation.

Alternatively, perhaps there is a better mathematical approach to find \( p \) and \( m \).

Let's consider that \( n^4 \equiv -1 \pmod{p^2} \), which implies that \( n^8 \equiv 1 \pmod{p^2} \). So, the order of \( n \) modulo \( p^2 \) divides 8.

As earlier, the order must be 8 because smaller exponents wouldn't satisfy \( n^4 \equiv -1 \).

The multiplicative group modulo \( p^2 \) has order \( p(p-1) \), so 8 must divide \( p(p-1) \).

Since \( p \) is prime, and greater than 2, \( p-1 \) is even, and for 8 to divide \( p(p-1) \), either:

- \( p \equiv 1 \pmod{8} \), or

- \( p \equiv 3 \pmod{4} \) and \( p-1 \) is divisible by 4, but 8 divides \( p(p-1) \), which would require \( p-1 \) to be divisible by 8.

Wait, actually, since \( p \) is odd (as \( p = 2 \) doesn't work), \( p-1 \) is even, and for 8 to divide \( p(p-1) \), since \( p \) is odd, \( p-1 \) must be divisible by 8.

Therefore, \( p \equiv 1 \pmod{8} \).

So, the smallest primes \( p \) to check are 17, 41, 73, etc.

Earlier, the code suggested \( p = 17 \), but the verification seems off.

Let me try to compute \( m = 110 \), \( p = 17 \), \( p^2 = 289 \), and check \( 110^4 + 1 \) modulo 289.

Compute \( 110 \mod 17 \):

\( 110 \div 17 = 6 \times 17 = 102 \), remainder 8, so \( 110 \equiv 8 \pmod{17} \).

Wait, but we need modulo 289.

Alternatively, perhaps I can compute \( 110^4 + 1 \) modulo 289 directly.

Let's compute \( 110^2 \mod 289 \):

\( 110^2 = 12100 \).

\( 12100 \div 289 \):

289 × 41 = 11849.

12100 - 11849 = 251.

So, \( 110^2 \equiv 251 \pmod{289} \).

Now, \( 110^4 = (110^2)^2 \equiv 251^2 \pmod{289} \).

Compute \( 251^2 \):

\( 251^2 = 63001 \).

Now, \( 63001 \div 289 \):

289 × 217 = 289 × 200 = 57,800; 289 × 17 = 4,913; total 57,800 + 4,913 = 62,713.

63,001 - 62,713 = 288.

So, \( 251^2 \equiv 288 \pmod{289} \).

Therefore, \( 110^4 \equiv 288 \pmod{289} \), so \( 110^4 + 1 \equiv 288 + 1 = 289 \equiv 0 \pmod{289} \).

So, indeed, \( m = 110 \) is a solution for \( p = 17 \).

Therefore, the least prime \( p \) is 17, and the least positive integer \( m \) such that \( m^4 + 1 \) is divisible by \( 17^2 = 289 \) is 110.

**Final Answer**

\[
\boxed{110}
\]

\end{tcolorbox}

\newpage
\subsubsection{A case form AIME25}
\begin{tcolorbox}[
colback=white,
colframe=black,
title=A case form AIME25,
breakable
]
\textcolor{red}{\textit{\textbf{Problem}:}}\\

The set of points in 3-dimensional coordinate space that lie in the plane \( x+y+z=75 \) whose coordinates satisfy the inequalities \( x-yz < y-zx < z-xy \) forms three disjoint convex regions. Exactly one of those regions has finite area. The area of this finite region can be expressed in the form \( a\sqrt{b} \), where \( a \) and \( b \) are positive integers and \( b \) is not divisible by the square of any prime. Find \( a+b \).

\textcolor{red}{\textit{\textbf{Solution}:}}\\
So I've got this problem here. It's about finding the area of a specific region in 3-dimensional space defined by some equations and inequalities. Let me try to wrap my head around it.

First, there's a plane given by the equation \(x + y + z = 75\). So, all the points we're considering lie on this plane. Then, there are these inequalities: \(x - yz < y - zx < z - xy\). The problem says that these conditions define three disjoint convex regions on the plane, and exactly one of them has finite area. We need to find the area of that finite region, and it's expressed in the form \(a\sqrt{b}\), where \(b\) is square-free, and then find \(a + b\).

Okay, let's start by understanding the inequalities. They are:

\[
x - yz < y - zx < z - xy
\]

This is a chain of inequalities, so actually, there are two inequalities here:

1. \(x - yz < y - zx\)

2. \(y - zx < z - xy\)

I need to manipulate these inequalities to get a better understanding of the regions they define.

Let's work on the first inequality:

\[
x - yz < y - zx
\]

Let's bring all terms to one side:

\[
x - yz - y + zx < 0
\]

Factor terms:

\[
x(1 + z) - y(1 + z) < 0
\]

Factor out \((1 + z)\):

\[
(x - y)(1 + z) < 0
\]

So, this inequality implies that \((x - y)(1 + z) < 0\).

Now, the second inequality:

\[
y - zx < z - xy
\]

Bring all terms to one side:

\[
y - zx - z + xy < 0
\]

Factor terms:

\[
y(1 + x) - z(1 + x) < 0
\]

Factor out \((1 + x)\):

\[
(y - z)(1 + x) < 0
\]

So, the second inequality is \((y - z)(1 + x) < 0\).

Now, combining both inequalities, we have:

1. \((x - y)(1 + z) < 0\)

2. \((y - z)(1 + x) < 0\)

These are the conditions that, along with the plane equation \(x + y + z = 75\), define the regions in question.

Next, I need to consider the plane \(x + y + z = 75\). Since we're dealing with 3D space, but constrained to this plane, it's essentially a 2D surface. To find the area of a region on this plane, I might need to parameterize the plane or find a way to project it onto a 2D coordinate system.

But before that, I need to understand how these inequalities divide the plane into regions.

Let's consider the equalities corresponding to the inequalities to find the boundaries:

1. \((x - y)(1 + z) = 0\)

2. \((y - z)(1 + x) = 0\)

These equalities hold when either \(x = y\) or \(z = -1\), and \(y = z\) or \(x = -1\).

But since we're dealing with a plane where \(x + y + z = 75\), and presumably \(x, y, z\) are real numbers (unless specified otherwise), I need to see how these boundaries intersect with the plane.

Wait a minute, maybe it's better to consider substituting \(z = 75 - x - y\) into the inequalities to express everything in terms of \(x\) and \(y\), reducing the dimensionality.

Let's try that.

Given \(z = 75 - x - y\), substitute into the inequalities.

First inequality:

\[
(x - y)(1 + z) < 0
\]

Substitute \(z\):

\[
(x - y)(1 + 75 - x - y) < 0
\]

Simplify:

\[
(x - y)(76 - x - y) < 0
\]

Second inequality:

\[
(y - z)(1 + x) < 0
\]

Substitute \(z\):

\[
(y - (75 - x - y))(1 + x) < 0
\]

Simplify:

\[
(y - 75 + x + y)(1 + x) < 0
\]

Combine like terms:

\[
(x + 2y - 75)(1 + x) < 0
\]

So now, the inequalities in terms of \(x\) and \(y\) are:

1. \((x - y)(76 - x - y) < 0\)

2. \((x + 2y - 75)(1 + x) < 0\)

\colorbox{yellow}{\parbox{\textwidth}{\centering \textbf{More reasoning content are omitted here}}}

Wait, using Python, I can experiment with a new method to tackle this challenge.

\begin{tcolorbox}[colback=cyan!5!white, colframe=cyan!75!black, title=Python Code]
\begin{verbatim}
import sympy as sp

# Define variables
x, y, z = sp.symbols('x y z')

# Plane equation
plane_eq = x + y + z - 75

# Inequalities
ineq1 = x - y*z < y - z*x
ineq2 = y - z*x < z - x*y

# Solve plane equation for z
z_expr = sp.solve(plane_eq, z)[0]

# Substitute z in inequalities
ineq1_sub = ineq1.subs(z, z_expr)
ineq2_sub = ineq2.subs(z, z_expr)

# Simplify inequalities
ineq1_simplified = sp.simplify(ineq1_sub)
ineq2_simplified = sp.simplify(ineq2_sub)

print((ineq1_simplified, ineq2_simplified))
\end{verbatim}
\end{tcolorbox}

\begin{tcolorbox}[colback=orange!5!white, colframe=orange!75!black, title=Output]
\begin{verbatim}
(x + y*(x + y - 75) < x*(x + y - 75) + y, x*y + x + y - 75 < -x*(x + y - 75) - y)
\end{verbatim}
\end{tcolorbox}

\colorbox{yellow}{\parbox{\textwidth}{\centering \textbf{More reasoning content are omitted here}}}

To find the area of this triangle, we can use the formula for the area of a triangle given by three points in 3D space:

\[
\text{Area} = \frac{1}{2} \| \overrightarrow{BC} \times \overrightarrow{BF} \|
\]

First, let's find the vectors \( \overrightarrow{BC} \) and \( \overrightarrow{BF} \):

\[
\overrightarrow{BC} = C - B = (-1 - 25, -1 - 25, 77 - 25) = (-26, -26, 52)
\]

\[
\overrightarrow{BF} = F - B = (-1 - 25, 38 - 25, 38 - 25) = (-26, 13, 13)
\]

Next, compute the cross product \( \overrightarrow{BC} \times \overrightarrow{BF} \).

Wait, I can use Python to perform complex calculations for this problem.
\begin{tcolorbox}[colback=cyan!5!white, colframe=cyan!75!black, title=Python Code]
\begin{verbatim}
import sympy as sp

# Define the symbols for the components of the vectors
x1, y1, z1 = -26, -26, 52  # Components of vector BC
x2, y2, z2 = -26, 13, 13   # Components of vector BF

# Create symbolic vectors
BC = sp.Matrix([x1, y1, z1])
BF = sp.Matrix([x2, y2, z2])

# Compute the cross product BC x BF
cross_product = BC.cross(BF)

# Print the cross product
print("Cross product BC x BF:", cross_product)

# Calculate the magnitude of the cross product
magnitude = sp.sqrt(cross_product.dot(cross_product))

print(magnitude)
\end{verbatim}
\end{tcolorbox}

\begin{tcolorbox}[colback=orange!5!white, colframe=orange!75!black, title=Output]
\begin{verbatim}
1014*sqrt(3)
\end{verbatim}
\end{tcolorbox}

Therefore, the area of the parallelogram is \( 1014 \sqrt{3} \), and the area of the triangle is half of that:

\[
\text{Area} = \frac{1}{2} \times 1014 \sqrt{3} = 507 \sqrt{3}
\]

Hence, \( a = 507 \) and \( b = 3 \), so \( a + b = 510 \).

**Final Answer**

\[
\boxed{510}
\]

\end{tcolorbox}

\newpage
\subsubsection{A case form MATH500}
\begin{tcolorbox}[
colback=white,
colframe=black,
title=A case form MATH500,
breakable
]
\textcolor{red}{\textit{\textbf{Problem}:}}\\
The function \( f \) satisfies the functional equation
\[
f(x) + f(y) = f(x + y) - xy - 1
\]
for all real numbers \( x \) and \( y \). If \( f(1) = 1 \), then find all integers \( n \) such that \( f(n) = n \). Enter all such integers, separated by commas.

\textcolor{red}{\textit{\textbf{Solution}:}}\\

So I have this functional equation: \( f(x) + f(y) = f(x + y) - xy - 1 \) for all real numbers \( x \) and \( y \). And I'm given that \( f(1) = 1 \). I need to find all integers \( n \) such that \( f(n) = n \).

First, I should probably try to find a general expression for \( f(x) \). To do that, I can start by plugging in some specific values for \( x \) and \( y \) to see if I can find a pattern or determine the form of \( f \).

Let me try setting \( y = 0 \). Then the equation becomes:

\[
f(x) + f(0) = f(x + 0) - x \cdot 0 - 1
\]

Which simplifies to:

\[
f(x) + f(0) = f(x) - 0 - 1
\]

So:

\[
f(0) = -1
\]

That's helpful. Now I know that \( f(0) = -1 \).

Next, maybe I can try setting \( y = 1 \), since I know \( f(1) = 1 \). So:

\[
f(x) + f(1) = f(x + 1) - x \cdot 1 - 1
\]

Substituting \( f(1) = 1 \):

\[
f(x) + 1 = f(x + 1) - x - 1
\]

Let me solve for \( f(x + 1) \):

\[
f(x + 1) = f(x) + 1 + x + 1 = f(x) + x + 2
\]

So, I have a recursive relation: \( f(x + 1) = f(x) + x + 2 \).

This seems like it could be useful for finding a pattern or for induction.

Let me try to find \( f(2) \), \( f(3) \), and so on, to see if I can find a pattern.

Starting with \( x = 1 \):

\[
f(2) = f(1) + 1 + 2 = 1 + 1 + 2 = 4
\]

Now \( x = 2 \):

\[
f(3) = f(2) + 2 + 2 = 4 + 2 + 2 = 8
\]

Wait, that doesn't seem right. Let me check the recursive formula again.

I have \( f(x + 1) = f(x) + x + 2 \).

So, for \( x = 1 \):

\[
f(2) = f(1) + 1 + 2 = 1 + 1 + 2 = 4
\]

For \( x = 2 \):

\[
f(3) = f(2) + 2 + 2 = 4 + 2 + 2 = 8
\]

Wait, but if I continue this pattern, \( f(3) = 8 \), then for \( x = 3 \):

\[
f(4) = f(3) + 3 + 2 = 8 + 3 + 2 = 13
\]

Hmm, I'm not seeing an immediate pattern here. Maybe there's a better way to approach this.

Alternatively, perhaps I can assume that \( f(x) \) is a quadratic function, since the functional equation involves a term like \( -xy \), which is linear in both variables.

Let me assume that \( f(x) = ax^2 + bx + c \), and try to find the coefficients \( a \), \( b \), and \( c \).

Substituting into the functional equation:

\[
f(x) + f(y) = f(x + y) - xy - 1
\]

So:

\[
(ax^2 + bx + c) + (ay^2 + by + c) = a(x + y)^2 + b(x + y) + c - xy - 1
\]

Let me expand both sides.

Left side:

\[
ax^2 + ay^2 + bx + by + 2c
\]

Right side:

\[
a(x^2 + 2xy + y^2) + b(x + y) + c - xy - 1 = ax^2 + 2axy + ay^2 + bx + by + c - xy - 1
\]

Now, set the left side equal to the right side:

\[
ax^2 + ay^2 + bx + by + 2c = ax^2 + 2axy + ay^2 + bx + by + c - xy - 1
\]

Subtract \( ax^2 + ay^2 + bx + by \) from both sides:

\[
2c = 2axy + c - xy - 1
\]

Wait, that seems off. Let me double-check the expansion.

Actually, perhaps Python can assist in managing these detailed computations automatically.

\begin{tcolorbox}[colback=cyan!5!white, colframe=cyan!75!black, title=Python Code]
\begin{verbatim}
from sympy import symbols, Eq, solve

# Define symbols
a, b, c, x, y = symbols('a b c x y')

# Define f(x), f(y), and f(x + y)
f_x = a*x**2 + b*x + c
f_y = a*y**2 + b*y + c
f_x_y = a*(x + y)**2 + b*(x + y) + c

# Define the functional equation
functional_eq = Eq(f_x + f_y, f_x_y - x*y - 1)

# Expand both sides
lhs = f_x + f_y
rhs = f_x_y - x*y - 1

# Simplify the equation
simplified_eq = Eq(lhs - rhs, 0)

# Collect like terms
collected_eq = simplified_eq.expand()

# Solve for coefficients a, b, c
solution = solve(collected_eq, (a, b, c))
print(solution)
\end{verbatim}
\end{tcolorbox}

\begin{tcolorbox}[colback=orange!5!white, colframe=orange!75!black, title=Output]
\begin{verbatim}
[(1/2, b, -1)]
\end{verbatim}
\end{tcolorbox}

From the symbolic computation, the solution for the coefficients is:

\[
a = \frac{1}{2}, \quad b = b, \quad c = -1
\]

Wait, it says \( b = b \), which means \( b \) can be any value. That suggests that there's some flexibility in choosing \( b \), but likely, there's another condition that can help determine \( b \).

Given that \( f(1) = 1 \), I can use this to find \( b \).

Given \( f(x) = \frac{1}{2}x^2 + b x - 1 \), and \( f(1) = 1 \):

\[
f(1) = \frac{1}{2}(1)^2 + b(1) - 1 = \frac{1}{2} + b - 1 = b - \frac{1}{2} = 1
\]

So:

\[
b - \frac{1}{2} = 1 \implies b = 1 + \frac{1}{2} = \frac{3}{2}
\]

Therefore, the function is:

\[
f(x) = \frac{1}{2}x^2 + \frac{3}{2}x - 1
\]

Now, I need to verify if this function satisfies the original functional equation.

Let's plug \( f(x) = \frac{1}{2}x^2 + \frac{3}{2}x - 1 \) and \( f(y) = \frac{1}{2}y^2 + \frac{3}{2}y - 1 \) into the left side:

\[
f(x) + f(y) = \left( \frac{1}{2}x^2 + \frac{3}{2}x - 1 \right) + \left( \frac{1}{2}y^2 + \frac{3}{2}y - 1 \right) = \frac{1}{2}x^2 + \frac{1}{2}y^2 + \frac{3}{2}x + \frac{3}{2}y - 2
\]

Now, the right side is:

\[
f(x + y) - xy - 1 = \left( \frac{1}{2}(x + y)^2 + \frac{3}{2}(x + y) - 1 \right) - xy - 1
\]

Expand \( f(x + y) \):

\[
\frac{1}{2}(x^2 + 2xy + y^2) + \frac{3}{2}x + \frac{3}{2}y - 1 = \frac{1}{2}x^2 + xy + \frac{1}{2}y^2 + \frac{3}{2}x + \frac{3}{2}y - 1
\]

\colorbox{yellow}{\parbox{\textwidth}{\centering \textbf{More reasoning content are omitted here}}}

Now, I need to find all integers \( n \) such that \( f(n) = n \).

Set \( f(n) = n \):

\[
\frac{1}{2}n^2 + \frac{3}{2}n - 1 = n
\]

Subtract \( n \) from both sides:

\[
\frac{1}{2}n^2 + \frac{3}{2}n - 1 - n = 0
\]

Simplify:

\[
\frac{1}{2}n^2 + \left( \frac{3}{2} - 1 \right)n - 1 = 0
\]

\[
\frac{1}{2}n^2 + \frac{1}{2}n - 1 = 0
\]

Multiply both sides by 2 to eliminate fractions:

\[
n^2 + n - 2 = 0
\]

Now, solve for \( n \):

\[
n = \frac{ -1 \pm \sqrt{1 + 8} }{2} = \frac{ -1 \pm 3 }{2}
\]

So, two solutions:

\[
n = \frac{ -1 + 3 }{2} = 1
\]

\[
n = \frac{ -1 - 3 }{2} = -2
\]

Therefore, the integers \( n \) that satisfy \( f(n) = n \) are \( n = 1 \) and \( n = -2 \).

**Final Answer**

\[
\boxed{-2, 1}
\]

\end{tcolorbox}

\newpage
\subsubsection{A case form GPQA}
\begin{tcolorbox}[
colback=white,
colframe=black,
title=A case form GPQA,
breakable
]
\textcolor{red}{\textit{\textbf{Problem}:}}\\
Astronomers are studying a star with a \( 1.5 \) solar radius and \( 1.1 \) solar masses. When the star's surface is not covered by dark spots, its \( T_{\text{eff}} \) is \( 6000 \, \text{K} \). However, when \( 40\% \) of its surface is covered by spots, the overall photospheric effective temperature decreases to \( 5500 \, \text{K} \). 

In the stellar photosphere, when examining the ratio of the number of neutral atoms of Ti in two energetic levels (level 1 and level 2), astronomers have observed that this ratio decreases when the star has spots. What is the factor by which this ratio changes when the star does not have spots compared to when it has spots? Note that the transition between the energy levels under consideration corresponds to a wavelength of approximately \( 1448 \, \text{Å} \). Assume that the stellar photosphere is in LTE.

Answer Choices: 
(A) \( \sim 2.9 \) \\
(B) \( \sim 4.5 \) \\
(C) \( \sim 1.1 \) \\
(D) \( \sim 7.8 \)

Output your final answer in boxes, such as \( \boxed{A} \).

% 
% \end{tcolorbox}

\textcolor{red}{\textit{\textbf{Solution}:}}\\

So I'm trying to solve this astronomy problem. It's about a star with some specific properties, and it's talking about how the temperature changes when there are spots on its surface. Then, it's about how the ratio of neutral titanium atoms in two different energy levels changes when there are spots. It's a bit complicated, but I'll try to break it down step by step.

First, let's understand what's given:

- The star has a \( 1.5 \) solar radius and \( 1.1 \) solar masses.
- When there are no spots, its effective temperature (\( T_{\text{eff}} \)) is \( 6000 \, \text{K} \).
- When \( 40\% \) of its surface is covered by spots, the overall \( T_{\text{eff}} \) decreases to \( 5500 \, \text{K} \).
- We need to find the factor by which the ratio of neutral titanium atoms in two energy levels changes when the star has spots compared to when it doesn't.
- The transition between the levels corresponds to a wavelength of about \( 1448 \, \text{Å} \).
- We're to assume that the stellar photosphere is in local thermodynamic equilibrium (LTE).

Okay, so the key here is to relate the change in effective temperature due to spots to the change in the ratio of titanium atoms in different energy levels.

First, I need to recall what effective temperature is. \( T_{\text{eff}} \) is a measure of the star's surface temperature, and it's related to the star's luminosity and radius through the Stefan-Boltzmann law:

\[
L = 4\pi R^2 \sigma T_{\text{eff}}^4
\]

Where \( L \) is luminosity, \( R \) is radius, and \( \sigma \) is the Stefan-Boltzmann constant.

But in this problem, the star's radius is given, and the masses are given, but I'm not sure if I need to use that information directly. Maybe it's just to provide context.

The main change is in \( T_{\text{eff}} \), from \( 6000 \, \text{K} \) to \( 5500 \, \text{K} \) when \( 40\% \) of the surface is covered by spots.

I need to find out how this change in \( T_{\text{eff}} \) affects the ratio of titanium atoms in two energy levels.

First, I should recall that in LTE, the ratio of atoms in two energy levels is given by the Boltzmann equation:

\[
\frac{N_2}{N_1} = \frac{g_2}{g_1} \exp\left(-\frac{\Delta E}{kT}\right)
\]

Where:

- \( N_2 \) and \( N_1 \) are the number densities of atoms in energy levels 2 and 1, respectively.
- \( g_2 \) and \( g_1 \) are the statistical weights of the two levels.
- \( \Delta E \) is the energy difference between the two levels.
- \( k \) is the Boltzmann constant.
- \( T \) is the temperature.

In this case, since we're dealing with the same element (titanium), and assuming that the statistical weights don't change, the ratio simplifies to:

\[
\frac{N_2}{N_1} \propto \exp\left(-\frac{\Delta E}{kT}\right)
\]

So, the ratio depends exponentially on the temperature.

Now, the problem states that the ratio decreases when the star has spots. That makes sense because spots are cooler regions on the star's surface, so the overall \( T_{\text{eff}} \) decreases, which would lead to a lower ratio according to the equation above.

But the question is asking for the factor by which the ratio changes when the star does not have spots compared to when it has spots.

So, I need to find the ratio of 

\[
\frac{N_2}{N_1} \bigg|_{\text{no spots}} \text{ to } \frac{N_2}{N_1} \bigg|_{\text{with spots}}.
\]

Let's denote:

\[
R_{\text{no spots}} = \frac{N_2}{N_1} \text{ at } T_{\text{eff}} = 6000 \, \text{K}
\]

\[
R_{\text{with spots}} = \frac{N_2}{N_1} \text{ at } T_{\text{eff}} = 5500 \, \text{K}
\]

\colorbox{yellow}{\parbox{\textwidth}{\centering \textbf{More reasoning content are omitted here}}}

But that seems too large. Maybe using Python here is a good idea.

\begin{tcolorbox}[colback=cyan!5!white, colframe=cyan!75!black, title=Python Code]
\begin{verbatim}
\begin{verbatim}
import math
# Given values
T_{\text{eff}} = 5500  \, \text{K}
T_u = 6000    \, \text{K}
\text{fraction\_spots} = 0.4
\text{fraction\_unspotted} = 0.6
# Calculate T_{\text{eff}}^4
T_{\text{eff\_4}} = T_{\text{eff}}^{4}
# Calculate T_u^4
T_{u\_4} = T_u^{4}
# Calculate T_s^4
T_{s\_4} = \frac{T_{\text{eff\_4}} - \text{fraction\_unspotted} \cdot 
T_{u\_4}}{\text{fraction\_spots}}
# Calculate T_s
T_s = T_{s\_4}^{1/4}
print(T_s)
\end{verbatim}

\begin{tcolorbox}[colback=orange!5!white, colframe=orange!75!black, title=Output]
\begin{verbatim}
4305.574037901423
\end{verbatim}
\end{tcolorbox}
\end{tcolorbox}

Wait, according to the calculation, \( T_s \approx 4305.57 \, \text{K} \), which is lower than the unspotted temperature of \( 6000 \, \text{K} \), as expected for spots.

I must have made a mistake in my earlier manual calculation. So, the temperature of the spots is approximately \( 4305.57 \, \text{K} \).

Given that, and considering that the answer choices are discrete, option C) \( \sim 1.1 \) is the closest.

Alternatively, perhaps there is a miscalculation in \( \Delta E / k \).

Let me double-check that with Python code.

\begin{tcolorbox}[colback=cyan!5!white, colframe=cyan!75!black, title=Python Code]
\begin{verbatim}
import numpy as np

# Constants
h = 6.626e-34  # Planck's constant in Joules second
c = 3.00e8     # Speed of light in meters per second
wavelength = 1.448e-7  # Wavelength in meters
k = 1.38e-23   # Boltzmann constant in Joules per Kelvin

# Calculate the energy difference \Delta E
delta_E = (h * c) / wavelength
print(f"\\Delta E = {delta_E:.2e} J")  # Output energy in scientific notation

# Calculate the ratio \Delta E / k
delta_E_over_k = delta_E / k
print(f"\\Delta E / k = {delta_E_over_k:.2f} K")  # Output in Kelvin

# Calculate the final factor
T1 = 5500  # Temperature in Kelvin
T2 = 6000  # Temperature in Kelvin
factor = np.exp(delta_E_over_k * (1/T1 - 1/T2))
print(f"Factor = {factor:.3f}")  # Output the factor
\end{verbatim}
\end{tcolorbox}

\begin{tcolorbox}[colback=orange!5!white, colframe=orange!75!black, title=Output]

$\Delta E = 1.37 \times 10^{-18} \, \text{J} $

$\frac{\Delta E}{k} = 99477.54 \, \text{K} $

$\text{Factor} = 4.514$

\end{tcolorbox}

Given that, and considering the answer choices, option B) \( \sim 4.5 \) is the closest. 

\textbf{Final Answer}

\[
\boxed{B}
\]

\end{tcolorbox}

\newpage
\subsubsection{A case form LiveCodeBench}
\begin{tcolorbox}[
colback=white,
colframe=black,
title=A case form LiveCodeBench,
breakable
]
\textcolor{red}{\textit{\textbf{Problem}:}}\\
You are given an array of positive integers \texttt{nums} of length $n$. \\
We call a pair of non-negative integer arrays $(arr1, arr2)$ monotonic if:\\
\begin{itemize}
    \item The lengths of both arrays are $n$.
    \item $arr1$ is monotonically non-decreasing, in other words, $arr1[0] \leq arr1[1] \leq \ldots \leq arr1[n - 1]$.
    \item $arr2$ is monotonically non-increasing, in other words, $arr2[0] \geq arr2[1] \geq \ldots \geq arr2[n - 1]$.
    \item $arr1[i] + arr2[i] == nums[i]$ for all $0 \leq i \leq n - 1$.
\end{itemize}
Return the count of monotonic pairs. \\
Since the answer may be very large, return it modulo $10^9 + 7$. \\

\textbf{Example 1:} \\
\textbf{Input: } nums = [2,3,2] \\
\textbf{Output: } 4 \\
\textbf{Explanation:} \\
The good pairs are: \\
\begin{itemize}
    \item ([0, 1, 1], [2, 2, 1])
    \item ([0, 1, 2], [2, 2, 0])
    \item ([0, 2, 2], [2, 1, 0])
    \item ([1, 2, 2], [1, 1, 0])
\end{itemize}

\textbf{Example 2:} \\
\textbf{Input: } nums = [5,5,5,5] \\
\textbf{Output: } 126 \\

\textbf{Constraints:} \\
$1 \leq n == \texttt{nums.length} \leq 2000$ \\
$1 \leq \texttt{nums[i]} \leq 50$ \\

You will use the following starter code to write the solution to the problem and enclose your code within delimiters.\\
% \begin{tcolorbox}[colback=cyan!5!white, colframe=cyan!75!black, title=Python Code]
% class Solution:\\
%     def countOfPairs(self, nums: List[int]) -> int:
% \end{tcolorbox}
\begin{tcolorbox}[colback=cyan!5!white, colframe=cyan!75!black, title=Python Code]
\begin{verbatim}
class Solution:
    def countOfPairs(self, nums: List[int]) -> int:
\end{verbatim}
\end{tcolorbox}

\textcolor{red}{\textit{\textbf{Solution}:}}\\

Alright, I've got this problem to solve. Let's see what it's asking for. I'm given an array of positive integers called nums, and I need to find the count of monotonic pairs \( (\text{arr1}, \text{arr2}) \). The conditions are:

\begin{enumerate}
    \item Both \text{arr1} and \text{arr2} have the same length as \text{nums}.
    \item \text{arr1} is monotonically non-decreasing, meaning each element is less than or equal to the next one.
    \item \text{arr2} is monotonically non-increasing, meaning each element is greater than or equal to the next one.
    \item For every index \( i \), \( \text{arr1}[i] + \text{arr2}[i] \) equals \text{nums}[i].
\end{enumerate}

And finally, I need to return the count of such pairs, modulo \( 10^9 + 7 \) because the number can be really big.

Okay, first things first, I need to understand what constitutes a valid pair \( (\text{arr1}, \text{arr2}) \). Given that \text{arr1} is non-decreasing and \text{arr2} is non-increasing, and their sum at each position equals \text{nums}[i], there must be some relationship between the possible values in \text{arr1} and \text{arr2}.

Let me consider a simple example to get a feel for the problem. Take \( \text{nums} = [2, 3, 2] \), which is in the example. The output is 4, and they provided four pairs that satisfy the conditions.

Let's list them out again:

\begin{enumerate}
    \item \(( [0, 1, 1], [2, 2, 1] )\)
    \item \(( [0, 1, 2], [2, 2, 0] )\)
    \item \(( [0, 2, 2], [2, 1, 0] )\)
    \item \(( [1, 2, 2], [1, 1, 0] )\)
\end{enumerate}

I can see that in each pair, \text{arr1} is non-decreasing, \text{arr2} is non-increasing, and their sums equal the corresponding \text{nums}[i].

Now, how can I generalize this to any given nums array?

I need to find all possible \text{arr1} and \text{arr2} that satisfy these properties.

One approach that comes to mind is to iterate through all possible \text{arr1} arrays that are non-decreasing, and for each \text{arr1}, determine if the corresponding \text{arr2} array, which is \text{nums} - \text{arr1}, is non-increasing.

But this seems inefficient, especially since \( n \) can be up to 2000, and the numbers can be up to 50. That means \text{nums}[i] can be up to 50, so \text{arr1}[i] and \text{arr2}[i] can range from 0 to 50.

Wait, but \text{arr1} and \text{arr2} are arrays of non-negative integers, and their sum is \text{nums}[i], which is at least 1 according to the constraints. So \text{arr1}[i] can range from 0 to \text{nums}[i], and \text{arr2}[i] would be \text{nums}[i] - \text{arr1}[i].

Given that, for each index \( i \), \text{arr1}[i] can be any integer from 0 to \text{nums}[i], and \text{arr2}[i] is determined once \text{arr1}[i] is chosen.

But I also have the constraints on the monotonicity of \text{arr1} and \text{arr2}.

So, \text{arr1} has to be non-decreasing, and \text{arr2} has to be non-increasing.

This means that for \text{arr1}, \( \text{arr1}[i] \leq \text{arr1}[i+1] \), and for \text{arr2}, \( \text{arr2}[i] \geq \text{arr2}[i+1] \).

Given that \( \text{arr2}[i] = \text{nums}[i] - \text{arr1}[i] \), I can substitute this into the condition for \text{arr2}.

So, 
\[
\text{nums}[i] - \text{arr1}[i] \geq \text{nums}[i+1] - \text{arr1}[i+1].
\]

Simplifying this, 
\[
\text{arr1}[i+1] - \text{arr1}[i] \geq \text{nums}[i+1] - \text{nums}[i].
\]

Let me denote \( d[i] = \text{nums}[i+1] - \text{nums}[i] \).

Then, 
\[
\text{arr1}[i+1] - \text{arr1}[i] \geq d[i].
\]

Additionally, since \text{arr1} is non-decreasing, 
\[
\text{arr1}[i+1] - \text{arr1}[i] \geq 0.
\]

Combining these, 
\[
\text{arr1}[i+1] - \text{arr1}[i] \geq \max(0, d[i]).
\]

This is interesting. It means that the difference between consecutive elements in \text{arr1} has a lower bound based on the difference in \text{nums}.

So, to build \text{arr1}, I need to ensure that each step increases by at least \( \max(0, d[i]) \), and overall, \text{arr1} is non-decreasing.

Given that, perhaps I can model this as a series of choices for the differences between consecutive \text{arr1} elements.

Let me define 
\[
\text{diff}[i] = \text{arr1}[i+1] - \text{arr1}[i],
\]
and 
\[
\text{diff}[i] \geq \max(0, d[i]).
\]

Also, since \text{arr1} is non-decreasing, \( \text{diff}[i] \geq 0 \).

But actually, \( \text{diff}[i] \geq \max(0, d[i]) \) already ensures that \( \text{diff}[i] \geq 0 \) when \( d[i] \geq 0 \).

When \( d[i] < 0 \), \( \text{diff}[i] \geq 0 \) suffices because \text{arr1} is non-decreasing.

Wait, let's think about \( d[i] \) being negative.

If \( d[i] = \text{nums}[i+1] - \text{nums}[i] < 0 \), then 
\[
\text{arr1}[i+1] - \text{arr1}[i] \geq d[i],
\]
but since \( d[i] \) is negative, and \text{arr1} is non-decreasing, \( \text{arr1}[i+1] - \text{arr1}[i] \geq 0 \) is already stronger than \( \text{arr1}[i+1] - \text{arr1}[i] \geq d[i] \).

Therefore, for all \( i \), 
\[
\text{diff}[i] \geq \max(0, d[i]).
\]

\colorbox{yellow}{\parbox{\textwidth}{\centering \textbf{More reasoning content are omitted here}}}

To ensure that my code runs correctly, I need to embed all test case inputs directly into my code and print the corresponding output, following the sample structure below:

\textbf{Debug Code Template}
\begin{lstlisting}[language=Python, backgroundcolor=\color{lightgray}]
class Solution:
    def countOfPairs(self, nums: List[int]) -> int:
        pass  # Implement your logic here
solution = Solution()
# Test the example inputs
test_input1 = ...
test_input2 = ...
# Print output
print(solution.countOfPairs(test_input1))
print(solution.countOfPairs(test_input2))  # Check the output
\end{lstlisting}

This is a template, not for execution. I need to write code that processes the actual given sample inputs locally for the task. Alright, with this structure, I can write and execute my code in a Python compiler using real example inputs. By comparing the actual outputs with the expected outputs, I can initially assess the correctness of my code. If the outputs do not match, I can debug accordingly. Recall the test cases in the problem statement.

\textbf{Example 1:}
\begin{itemize}
    \item \textbf{Input:} nums = [2,3,2]
    \item \textbf{Output:} 4
    \item \textbf{Explanation:} The good pairs are:
    \begin{itemize}
        \item ([0, 1, 1], [2, 2, 1])
        \item ([0, 1, 2], [2, 2, 0])
        \item ([0, 2, 2], [2, 1, 0])
        \item ([1, 2, 2], [1, 1, 0])
    \end{itemize}
\end{itemize}

\textbf{Example 2:}
\begin{itemize}
    \item \textbf{Input:} nums = [5,5,5,5]
    \item \textbf{Output:} 126
\end{itemize}

Alright, now I can write a debug code with samples input.

\begin{tcolorbox}[colback=cyan!5!white, colframe=cyan!75!black, title=Python Code]
\begin{verbatim}
class Solution:
    def countOfPairs(self, nums: List[int]) -> int:
        MOD = 10**9 + 7
        n = len(nums)
        if n == 0:
            return 0
        max_num = max(nums)
        dp = [[0] * (max_num + 1) for _ in range(2)]
        for x in range(nums[0] + 1):
            dp[0][x] = 1
        for i in range(1, n):
            m = max(0, nums[i] - nums[i-1])
            dp[i % 2] = [0] * (max_num + 1)
            prefix = [0] * (max_num + 1)
            for x in range(max_num + 1):
                if x - m >= 0:
                    dp[i % 2][x] = dp[(i-1) % 2][x - m]
                else:
                    dp[i % 2][x] = 0
                prefix[x] = (prefix[x-1] + dp[i % 2][x]) % MOD if 
                x >= 1 else dp[i % 2][x]
            dp[(i-1) % 2] = prefix
        return sum(dp[(n-1) % 2][:nums[-1]+1]) % MOD

\end{verbatim}
\end{tcolorbox}
\begin{tcolorbox}[colback=cyan!5!white, colframe=cyan!75!black, title=Python Code]
\begin{verbatim}
# Test the example inputs
solution = Solution()

# Example input1
test_input1 = [2,3,2]
# Print output
print(solution.countOfPairs(test_input1))  # Expected Output: 4

# Example input2
test_input2 = [5,5,5,5]
# Print output
print(solution.countOfPairs(test_input2))  # Expected Output: 126
\end{verbatim}
\end{tcolorbox}
\begin{tcolorbox}[colback=orange!5!white, colframe=orange!75!black, title=Output]
\begin{verbatim}
2
6
\end{verbatim}
\end{tcolorbox}

It appears that the current implementation is not producing the correct outputs for the provided examples. According to the explanation, the first example should yield 4, but it's giving 2, and the second example should yield 126, but it's giving 6. This indicates a flaw in the logic or implementation.

Let's revisit the approach.

\textbf{Corrected Approach}
\begin{enumerate}
    \item \textbf{Initialization}:
    \begin{itemize}
        \item We need to initialize a DP table where $dp[i][x]$ represents the number of ways to form the first $i$ elements of $arr1$ such that $arr1[i-1] = x$.
    \end{itemize}

    \item \textbf{Transition}:
    \begin{itemize}
        \item For each position $i$ from 1 to $n-1$, and for each possible value $x$ that $arr1[i]$ can take (from 0 to $nums[i]$), we determine the valid range of $arr1[i-1]$ based on the constraints.
        \item Specifically, $arr1[i] \geq arr1[i-1]$ and $arr2[i] \leq arr2[i-1]$, which translates to $arr1[i] \geq arr1[i-1] + m[i]$, where $m[i] = \max(0, nums[i] - nums[i-1])$.
    \end{itemize}

    \item \textbf{Prefix Sums}:
    \begin{itemize}
        \item Use prefix sums to efficiently compute the sum of $dp[i-1][y]$ for $y$ in the valid range.
    \end{itemize}

    \item \textbf{Final Result}:
    \begin{itemize}
        \item Sum up $dp[n-1][x]$ for all $x$ from 0 to $nums[-1]$.
    \end{itemize}
\end{enumerate}

\colorbox{yellow}{\parbox{\textwidth}{\centering \textbf{More reasoning content are omitted here}}}

To fix the issues, we need to ensure that the prefix sums are correctly maintained and used in the DP transitions. Here is the revised code:

\begin{tcolorbox}[colback=cyan!5!white, colframe=cyan!75!black, title=Python Code]
\begin{verbatim}
from typing import List

class Solution:
    def countOfPairs(self, nums: List[int]) -> int:
        MOD = 10**9 + 7
        n = len(nums)
        if n == 0:
            return 0
        max_num = max(nums)
        dp = [[0] * (max_num + 1) for _ in range(2)]
        for x in range(nums[0] + 1):
            dp[0][x] = 1
        prefix = [0] * (max_num + 1)
        for x in range(max_num + 1):
            prefix[x] = dp[0][x] if x == 0 else (prefix[x-1] + dp[0][x]) % MOD
        for i in range(1, n):
            m = max(0, nums[i] - nums[i-1])
            current_row = i % 2
            dp[current_row] = [0] * (max_num + 1)
            new_prefix = [0] * (max_num + 1)
            for x in range(max_num + 1):
                if x - m >= 0:
                    dp[current_row][x] = prefix[x - m]
                else:
                    dp[current_row][x] = 0
                new_prefix[x] = (new_prefix[x-1] + dp[current_row][x]) % MOD if 
                x >= 1 else dp[current_row][x]
            prefix = new_prefix
        return sum(dp[(n-1) % 2][:nums[-1]+1]) % MOD

# Test the example inputs
solution = Solution()

# Example input1
test_input1 = [2,3,2]
# Print output
print(solution.countOfPairs(test_input1))  # Expected Output: 4

# Example input2
test_input2 = [5,5,5,5]
# Print output
print(solution.countOfPairs(test_input2))  # Expected Output: 126
\end{verbatim}
\end{tcolorbox}

\begin{tcolorbox}[colback=orange!5!white, colframe=orange!75!black, title=Output]
\begin{verbatim}
4
126
\end{verbatim}
\end{tcolorbox}

Great! The revised code now produces the correct outputs for the provided examples. This indicates that the approach and implementation are accurate.

\colorbox{yellow}{\parbox{\textwidth}{\centering \textbf{More reasoning content are omitted here}}}

\begin{tcolorbox}[colback=cyan!5!white, colframe=cyan!75!black, title=Python Code]
\begin{verbatim}
from typing import List

class Solution:
    def countOfPairs(self, nums: List[int]) -> int:
        MOD = 10**9 + 7
        n = len(nums)
        if n == 0:
            return 0
        max_num = max(nums)
        dp = [[0] * (max_num + 1) for _ in range(2)]
        for x in range(nums[0] + 1):
            dp[0][x] = 1
        prefix = [0] * (max_num + 1)
        for x in range(max_num + 1):
            prefix[x] = dp[0][x] if x == 0 else (prefix[x-1] + dp[0][x]) % MOD
        for i in range(1, n):
            m = max(0, nums[i] - nums[i-1])
            current_row = i % 2
            dp[current_row] = [0] * (max_num + 1)
            new_prefix = [0] * (max_num + 1)
            for x in range(max_num + 1):
                if x - m >= 0:
                    dp[current_row][x] = prefix[x - m]
                else:
                    dp[current_row][x] = 0
                new_prefix[x] = (new_prefix[x-1] + dp[current_row][x]) % MOD if
                x >= 1 else dp[current_row][x]
            prefix = new_prefix
        return sum(dp[(n-1) % 2][:nums[-1]+1]) % MOD
\end{verbatim}
\end{tcolorbox}

The final solution has been successfully implemented and validated with the provided test cases. This approach ensures that we efficiently count the number of monotonic pairs while handling the constraints and large input sizes appropriately.
\textbf{Explanation:}
- \textbf{Initialization:} We initialize a DP table for the first element of \texttt{nums}, setting possible values for \texttt{arr1[0]}.
- \textbf{Transition:} For each subsequent element, we calculate the minimum difference and use prefix sums to efficiently compute the number of valid sequences.
- \textbf{Final Calculation:} We sum up the valid sequences for the last element to get the total count of monotonic pairs.

This method ensures that we handle the constraints effectively and compute the result efficiently even for larger input sizes.

\end{tcolorbox}

\label{sec:appendix}

% This is an appendix.

\end{document}